\pdfoutput=1

\documentclass[11pt]{article}

\usepackage[final]{acl}
\usepackage{url}
\usepackage{times}
\usepackage{latexsym}
\usepackage{enumitem}
\usepackage[T1]{fontenc}

\usepackage[utf8]{inputenc}

\usepackage{microtype}

\usepackage{inconsolata}

\usepackage{graphicx}
\usepackage{times}
\usepackage{latexsym}

\usepackage[T1]{fontenc}
\usepackage[utf8]{inputenc}

\usepackage{microtype}

\usepackage{inconsolata}

\usepackage{graphicx}

\usepackage{enumitem}
\usepackage{svg}
\usepackage{booktabs}

\usepackage{colortbl}
\usepackage{pgfmath}
\usepackage[most]{tcolorbox}

\usepackage{url}

\usepackage[most]{tcolorbox} 
\tcbset{
  myboxstyle/.style={
    colback=gray!10!white,   
    colframe=gray!70!black,  
    fonttitle=\bfseries,     
    coltitle=white,          
    boxrule=0.3mm,           
    arc=2mm,                 
    left=1mm, right=1mm,     
    top=1mm, bottom=1mm      
  }
}
\usepackage{siunitx}         
\usepackage{pgf}             
\usepackage{ifthen}          

\usepackage{booktabs}
\usepackage{multirow}
\usepackage{todonotes}

\newcommand{\perc}[2]{%
  \pgfmathsetmacro{\increase}{((#2 - #1) / #1 * 100)}%
  \pgfmathparse{round(\increase)}%
  \pgfmathtruncatemacro{\increaseInt}{\pgfmathresult}%
  \pgfmathtruncatemacro{\displayPerc}{abs(\increaseInt)}%
  \pgfmathsetmacro{\colorlevel}{max(0, min(100, \displayPerc * 3))}%
  \ifdim \increase pt > 0pt
    \def\arrowSymbol{$\uparrow$}%
    \def\cellColor{green}%
  \else
    \def\arrowSymbol{$\downarrow$}%
    \def\cellColor{red}%
  \fi

  \edef\temp{\noexpand\cellcolor{\cellColor!\colorlevel} \num[round-precision=0]{#2}\, \arrowSymbol\displayPerc\%\ }%
  \temp%
}

\title{\textcolor{blue}{MAL}icious \textcolor{blue}{INT}ent Dataset and Inoculating LLMs for Enhanced Disinformation Detection}

\author{
 \textbf{Arkadiusz Modzelewski\textsuperscript{1,2,3}}, 
 \textbf{Witold Sosnowski\textsuperscript{2}},
 \textbf{Eleni Papadopulos\textsuperscript{1,4}},
 \textbf{Elisa Sartori\textsuperscript{1}},\\
 \textbf{Tiziano Labruna\textsuperscript{1}},
 \textbf{Giovanni Da San Martino\textsuperscript{1}},
 \textbf{Adam Wierzbicki\textsuperscript{2}}
 \vspace{0.2cm}
\\
 \textsuperscript{1}University of Padua, Italy \\
 \textsuperscript{2}Polish-Japanese Academy of Information Technology, Poland \\
 \textsuperscript{3}NASK National Research Institute, Poland \\
 \textsuperscript{4}{Politecnico di Torino}, Italy \\
 \small{
   \textbf{Correspondence:} \href{mailto:contact@amodzelewski.com}{contact@amodzelewski.com}
 }
}

\begin{document}
\maketitle
\begin{abstract}


The intentional creation and spread of disinformation poses a significant threat to public discourse. However, existing English datasets and research rarely address the intentionality behind the disinformation. This work presents MALINT, the first human-annotated English corpus developed in collaboration with expert fact-checkers to capture disinformation and its malicious intent. We utilize our novel corpus to benchmark 12 language models, including small language models (SLMs) such as BERT and large language models (LLMs) like Llama 3.3, on binary and multilabel intent classification tasks. Moreover, inspired by inoculation theory from psychology and communication studies, we investigate whether incorporating knowledge of malicious intent can improve disinformation detection. To this end, we propose intent-based inoculation, an intent‑augmented reasoning for LLMs that integrates intent analysis to mitigate the persuasive impact of disinformation. Analysis on six disinformation datasets, five LLMs, and seven languages shows that intent‑augmented reasoning improves zero‑shot disinformation detection. To support research in intent‑aware disinformation detection, we release the MALINT dataset with annotations from each annotation step.

\end{abstract}

\section{Introduction}
\label{sec:intro}

The creation, dissemination, and consumption of online disinformation are increasing concerns, driven by easy access to false content and limited public awareness of its misleading nature \cite{shu2020combating}. The High Level Expert Group, established by the European Commission, defines disinformation as “\textit{false, inaccurate, or misleading information designed, presented, and promoted to intentionally cause public harm or for profit}” \cite{de2018multi}. 
Researchers in communication theory further stress the importance of intentionality in defining disinformation \cite{hameleers2023disinformation}.
Uncovering intentions can help future research detect more effectively goal-driven attempts to influence public beliefs \cite{hameleers2023disinformation}. Although English resources support disinformation research \cite{wang2017liar, shu2020fakenewsnet, ahmed2018detecting}, none address the varying types of intent behind malicious agents. To fill this gap, we introduce \textbf{MALINT}, the first English corpus that annotates disinformation and the most common \textbf{MAL}icious \textbf{INT}ention types of disinformation agents. The MALINT  dataset is a high-quality resource developed in collaboration with fact-checking experts from organizations accredited by the International Fact-Checking Network (IFCN)\footnote{The IFCN accredits fact-checking and debunking organizations that adhere to its code of principles. See \url{ https://www.poynter.org/ifcn/}}. \mbox{We use MALINT to pursue two core objectives.} 
The first one is \textbf{Intent Classification}.
We present the first investigation into how well different language models (LMs) can detect malicious intent in English texts. We evaluate small and large language models on binary and multilabel classification tasks.
The second objective is \textbf{Intent-Augmented Disinformation Detection}.
Inoculation theory in psychology suggests that exposing individuals to weakened forms of disinformation can build resistance to deception \cite{traberg2022psychological, roozenbeek2020prebunking}. Building on this idea, we explore whether weakening disinformation via integration of malicious intent knowledge can enhance the LLMs' zero-shot disinformation detection. To evaluate it, we propose intent-based inoculation (IBI) and conduct experiments on five established disinformation datasets that include only disinformation labels, as well as on MALINT. In analysis, we use three data splits: (a) a genre-based (articles vs. posts), (b) a temporal split separating texts published before and after the LLMs' knowledge cutoff dates, and (c) a language split employed to assess the usefulness of intent-based reasoning in a multilingual context, covering even low-resource languages such as Estonian and Polish. We demonstrate that IBI outperforms competitive methods by an average of 9\% in English and achieves even larger gains in other languages. In summary, our main contributions are:
\begin{itemize}[nosep, leftmargin=*]
    \item Our MALINT, is the first English corpus annotated for \textit{malicious intent and disinformation} comprising comprehensive stepwise annotations.
    \item We evaluate malicious intent classification capabilities of 12 different language models.
    \item By leveraging IBI, 
    we show that intent reasoning improves LLM disinformation detection across diverse datasets and languages.
\end{itemize}
\noindent We release our dataset, prompts, and codebase\footnote{Repository with data, prompts and codebase: \url{https://github.com/ArkadiusDS/MALINT}}.

\section{The MALINT Dataset}
\label{sec:malint_dataset}

\textbf{MALINT} is a novel corpus of online news articles designed to advance research on disinformation and the malicious intents behind it. 
During annotation, annotators first assess each article’s credibility. Articles deemed disinformative are then annotated for the underlying malicious intent of the disinformation agents. This approach is grounded in the recognition that disinformation is deliberately crafted to serve specific malicious objectives \cite{hameleers2023disinformation}. Credible content, by its nature, is free of such intent. This perspective draws on the growing consensus in disinformation research that malicious intent is a defining feature of disinformative content \cite{Zhou_2022, appelman2022truth, hameleers2023disinformation, wang2024misinformation}. 

\subsection{Data Sources and Collection}\label{ssec:data_sources}

To build a representative dataset, we collected articles from about 50 online sources spanning mainstream media, outlets promoting alternative or incidental narratives. Sources were reviewed by fact-checking experts and classified by consensus into one of three categories: \textit{Reliable}, \textit{Unreliable}, or \textit{Mixed/Biased}. Classification was based on systematic content review and cross-checking with fact-checking tools such as Media Bias/Fact Check\footnote{An initiative where domain experts perform a careful manual analysis based on clear guidelines \cite{nakov2024survey} Link: \href{https://mediabiasfactcheck.com/}{https://mediabiasfactcheck.com/}}. Articles were collected from all sources and subsequently provided for disinformation and malicious intent annotation. To prevent annotation bias, source categories were hidden from annotators. We release the full list of sources.

\subsection{Annotation Methodology and Guidelines}

A rigorous, multi-stage annotation approach was used to ensure high-quality and consistent annotations of the dataset. The following outlines the creation of guidelines, annotator training, and the step-by-step workflow adopted for article labeling.

\paragraph{Guidelines Creation.}

Our project began with the development of detailed guidelines by a team of experienced disinformation researchers and fact-checking experts, each with 3+ years of expertise in IFCN-accredited organizations. The guidelines specified annotation categories and established rules for ambiguous or complex cases, ensuring a consistent and robust framework for the project (\mbox{Appendix \ref{appendix:guidelines}} presents annotation guidelines).

\paragraph{Annotator Training and Calibration.}
To ensure consistent application of guidelines, all annotators participated in a training that combined remote and on-site sessions. The training featured hands-on exercises and calibration annotation rounds, allowing annotators to converge in their understanding of guidelines and receive targeted feedback from the lead fact-checking trainer. Annotations from the calibration phase were used solely for training purposes and were excluded from the final dataset.

\paragraph{Annotation and Review Workflow.}
After training, annotation followed a structured workflow to ensure quality and reliability:
\begin{enumerate}[nosep, leftmargin=*]  
    \item \textbf{Independent Annotation}: Each article was independently labeled by a primary annotator and by a supervisor with expertise in disinformation. Discrepancies were resolved through discussion to reach a consensus. The supervisor also conducted a third annotation, with unresolved cases passed to the next stage if necessary.
    \item \textbf{Resolving Ambiguities}:  If the primary annotator and supervisor could not reach consensus, the article could be reviewed with a senior fact-checking expert. If it remained ambiguous after this step, it was labeled as \textit{Hard-to-say}.
\end{enumerate}

\paragraph{Credibility Annotation.}
Each article was reviewed using a methodology that incorporated the debunking technique, complemented by fact-checking principles, as outlined by the NATO Strategic Communications Centre of Excellence~\cite{pamment2021fact}. Annotators assigned one of three labels. Two main annotations are: \textit{Credible Information} and \textit{Disinformation}. We used the definition of disinformation proposed by the European Commission’s High‑Level Expert Group~\cite{de2018multi}, widely applied in recent research~\cite{hameleers2023disinformation, modzelewski2024mipd, sosnowski2024eu}. Moreover, we introduced an additional annotation label: \textit{Hard‑to‑say}, for cases where annotators could not agree on veracity.  Articles falling into the latter class were excluded from our experiments.

\paragraph{Malicious Intent Annotation.}
Given that disinformation is deliberately disseminated, we defined five intent categories: \textit{Undermining the Credibility of Public Institutions} (UCPI), \textit{Changing Political Views} (CPV), \textit{Undermining International Organizations and Alliances} (UIOA), \textit{Promoting Social Stereotypes/Antagonisms} (PSSA), and \textit{Promoting Anti-scientific Views} (PASV). Since annotators could assign any number of these categories to a single article (including none or all) this task constitutes a multilabel annotation problem.

\begin{figure}[ht!]
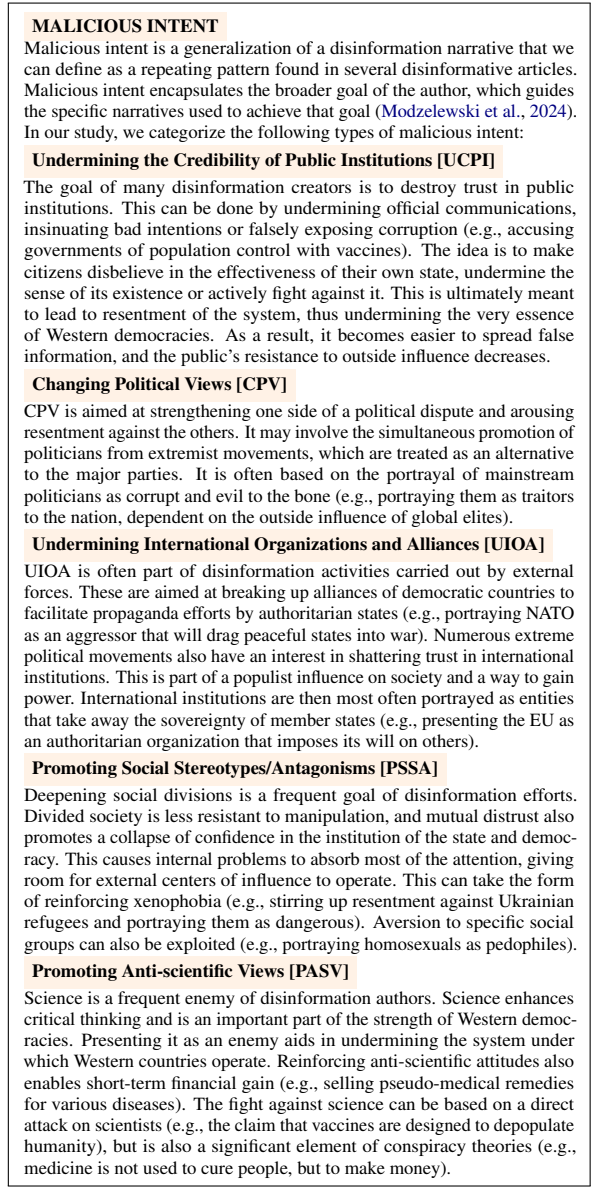

  \fbox{
    \begin{minipage}{0.94\columnwidth}
      \scriptsize 
      \colorbox{orange!10}{\textbf{MALICIOUS INTENT}} 
       \\ Malicious intent is a generalization of a disinformation narrative that we can define as a repeating pattern found in several disinformative articles. Malicious intent encapsulates the broader goal of the author, which guides the specific narratives used to achieve that goal \cite{modzelewski2024mipd}. In our study, we categorize the following types of malicious intent:  \\ 
      \colorbox{orange!10}{\textbf{Undermining the Credibility of Public Institutions [UCPI]}} \\ The goal of many disinformation creators is to destroy trust in public institutions. This can be done by undermining official communications, insinuating bad intentions or falsely exposing corruption (e.g., accusing governments of population control with vaccines). The idea is to make citizens disbelieve in the effectiveness of their own state, undermine the sense of its existence or actively fight against it. This is ultimately meant to lead to resentment of the system, thus undermining the very essence of Western democracies. As a result, it becomes easier to spread false information, and the public's resistance to outside influence decreases.
      
      \colorbox{orange!10}{\textbf{Changing Political Views [CPV]}} \\
      CPV is aimed at strengthening one side of a political dispute and arousing resentment against the others. It may involve the simultaneous promotion of politicians from extremist movements, which are treated as an alternative to the major parties. It is often based on the portrayal of mainstream politicians as corrupt and evil to the bone (e.g., portraying them as traitors to the nation, dependent on the outside influence of global elites).

      \colorbox{orange!10}{\textbf{Undermining International Organizations and Alliances [UIOA]}} \\      
        UIOA is often part of disinformation activities carried out by external forces. These are aimed at breaking up alliances of democratic countries to facilitate propaganda efforts by authoritarian states (e.g., portraying NATO as an aggressor that will drag peaceful states into war). Numerous extreme political movements also have an interest in shattering trust in international institutions. This is part of a populist influence on society and a way to gain power. International institutions are then most often portrayed as entities that take away the sovereignty of member states (e.g., presenting the EU as an authoritarian organization that imposes its will on others). 
      
      \colorbox{orange!10}{\textbf{Promoting Social Stereotypes/Antagonisms [PSSA]}} \\
        Deepening social divisions is a frequent goal of disinformation efforts. Divided society is less resistant to manipulation, and mutual distrust also promotes a collapse of confidence in the institution of the state and democracy. This causes internal problems to absorb most of the attention, giving room for external centers of influence to operate. This can take the form of reinforcing xenophobia (e.g., stirring up resentment against Ukrainian refugees and portraying them as dangerous). Aversion to specific social groups can also be exploited (e.g., portraying homosexuals as pedophiles).
      
      \colorbox{orange!10}{\textbf{Promoting Anti-scientific Views  [PASV]}} \\  
        Science is a frequent enemy of disinformation authors. Science enhances critical thinking and is an important part of the strength of Western democracies. Presenting it as an enemy aids in undermining the system under which Western countries operate. Reinforcing anti-scientific attitudes also enables short-term financial gain (e.g., selling pseudo-medical remedies for various diseases). The fight against science can be based on a direct attack on scientists (e.g., the claim that vaccines are designed to depopulate humanity), but is also a significant element of conspiracy theories (e.g., medicine is not used to cure people, but to make money).
    \end{minipage}
  }
  \caption{Definition and categories of malicious intent.}
  \label{fig:mal_intent_taxonomy}
\end{figure}

Our categories and intent definition are based on the study proposed by \citet{modzelewski2024mipd} and refined in collaboration with fact-checking experts to better reflect the current disinformation landscape. Figure~\ref{fig:mal_intent_taxonomy} shows malicious intent definition and detailed descriptions for each category. 


\subsection{Annotation and Data Quality Control}

To ensure annotation reliability and reduce bias, each article was independently reviewed by two annotators (a primary annotator and their supervisor). The supervisor performed a third pass, considering independent annotations. In the event of disagreement, supervisors were encouraged to consult with the initial annotators and, as needed, with a senior fact-checking expert.

In the first stage, two annotators achieved an agreement of approximately 85.31\% on the credibility task. They reached 65.19\% agreement on the more complex multilabel intent task. These figures reflect pre-consensus agreement and indicate how challenging it was to prepare the final consensus annotation rather than measuring the final label quality \cite{piskorski2023multilingual}.

In the second stage, the supervisor performed a third annotation. Disagreements were resolved through consensus, and expert input was utilized when necessary. At this stage, the annotation process concluded with agreement exceeding 95\% between supervisors for both tasks. This improved the reliability and quality of the dataset. If consensus could not be reached for credibility analysis, the article was assigned a \textit{Hard-to-say} label and excluded from further steps.

\noindent\textbf{Note}: We publish annotations from each stage.

\subsection{Dataset Statistics}\label{ssec:malint_stats}

Table~\ref{tab:malint_overview} presents the key statistics of the \textbf{MALINT} dataset, which consists of 1,600 news articles, providing a substantial corpus for analysis.

\paragraph{Credibility and Malicious Intent Labels.}
The dataset includes two primary credibility labels: \textit{Credible}, comprising 63.5\% instances, and \textit{Disinformation}, which accounts for the remaining 584 articles (36.5\%).
Table~\ref{tab:intent_distribution} details the distribution of the five malicious intent categories in dataset. 

\begin{table}[ht]
\centering
\scriptsize
\begin{tabular}{@{}l c@{}}
\toprule
\textbf{Statistic} & \textbf{Value} \\
\midrule
Total No. of Articles & 1{,}600 \\
Avg. Article Length (words) & 963 \\
Avg. Article Length (characters) & 6{,}045 \\
\bottomrule
\end{tabular}
\caption{Overview of the \textbf{MALINT} corpus.}
\label{tab:malint_overview}
\end{table}

\begin{table}[ht]
\centering
\scriptsize
\begin{tabular}{@{}l c c c c c@{}}
\toprule
\textbf{Statistic} & \textbf{UCPI} & \textbf{UIOA} &  \textbf{PASV} & \textbf{PSSA} & \textbf{CPV}  \\
\midrule
Count & 321 & 234 & 154 & 222 & 197 \\
\% & 20.06 & 14.63 & 9.63 & 13.88 & 12.31 \\
\bottomrule
\end{tabular}
\caption{Malicious intent types distribution in MALINT.}
\label{tab:intent_distribution}
\end{table}

\paragraph{Malicious Intents Multiplicity.}
Approximately 12{.}1\% of articles are tagged with a single intent, while around 24\% contain two or more intent labels. Among these, the most common pattern is the presence of exactly two intents, observed in 15{.}5\% of all articles. The most frequent intent pair is \textit{UIOA} and \textit{UCPI}, co-occurring in 127 articles.

\section{Intent Classification}
\label{sec:lms_evaluation}

To evaluate the ability of LMs to detect malicious intent, we use the MALINT dataset to assess performance across two classification tasks. These tasks are intended to capture different dimensions of intent recognition and provide a broad view of model accuracy when faced with malicious content. We evaluate LMs on the following tasks:

\begin{itemize}[nosep, leftmargin=*]
    \item \textbf{Binary Detection Per Class} -  Models are evaluated on their ability to detect the presence of a specific malicious intent. Each intent is treated as an independent binary classification problem. This setup allows us to analyze how well models can identify individual intent categories.
    
    \item \textbf{Multilabel Detection} - Evaluating a model's ability to simultaneously identify multiple intent types that may co-occur in a given input. The task is formulated as a multilabel classification problem, where models must assign all relevant intent labels to each input.
\end{itemize}
As shown in Table~\ref{tab:intent_distribution}, our tasks are challenging due to class imbalance across intent categories.

\subsection{Experimental Setup}

For our experiments, the MALINT dataset was split into 770 training, 330 validation, and 500 test instances. Binary classification was evaluated using F\textsubscript{1} over the positive class, while the multilabel task used weighted F\textsubscript{1} to address class imbalance. All metrics were computed on test sets.

\paragraph{Setup for SLMs.}

We fine-tuned a range of pre-trained SLMs, selected to represent different architectures and computational requirements. We have chosen \textit{BERT} \cite{DBLP:journals/corr/abs-1810-04805}, \textit{RoBERTa} \cite{DBLP:journals/corr/abs-1907-11692}, \textit{DeBERTa V3} \cite{he2021deberta, he2021debertav3} and \textit{DistilBERT} \cite{Sanh2019DistilBERTAD}. DistilBERT was included to assess the model suitable for environments with limited computational resources.
Each model was fine-tuned for the two tasks across 42 model-task combinations, testing multiple hyperparameter settings, totaling around 2{,}000 experiments.  All experiments were run on an NVIDIA L40 GPU.
Details and the optimal hyperparameters for each experiment are provided in Appendix~\ref{appendix:exp_benchmark_setup}.

\paragraph{Setup for LLMs.}

We evaluated five cutting-edge LLMs via different APIs: \textit{GPT 4o Mini},
\textit{GPT 4.1 Mini}, \textit{Gemini 2.0 Flash}, \textit{Gemma 3 27b it}, \textit{Llama 3.3 70B}. To ensure as deterministic results as possible, we prompted all models with the temperature parameter set to zero. All evaluations were conducted in a zero-shot setting, as many documents were too long for few-shot prompting within the LLM context limits. The full set of experiments involved approximately 15,000 API calls. Prompts for all tasks are provided in Appendix~\ref{appendix:prompts_malint}. 

Further details regarding APIs used and each model, including their knowledge cut-off dates and the rationale for their selection, are available in Appendix~\ref{appendix:llms_details}.



\paragraph{Baselines.}

We implemented two baselines for all tasks: a random classifier and logistic regression. For binary classification, the logistic regression model was trained on bag-of-words features represented as sparse token count vectors, with English stop words removed. For the multilabel task, logistic regression was applied using a one-vs-rest strategy \cite{murphy2018machine}.

\subsection{Evaluation Results}
\paragraph{Binary Detection Per Each Class.}

As shown in Table~\ref{tab:intent-per-class-slms-results}, DeBERTa V3 Large and RoBERTa models consistently achieved the highest F\textsubscript{1} scores across most intent categories among SLMs.
Among LLMs, \mbox{GPT 4.1} Mini performed best for \textit{UCPI} and \textit{PSSA}, while Llama 3.3 70B for \textit{PASV} and \textit{CPV} categories. LLMs achieved superior results compared to fine-tuned SLMs across three intent categories. 
The logistic regression baseline with bag-of-words outperformed random predictions but remained below most LMs, serving as a simple baseline.

\begin{table}[h]
\centering
\scriptsize
\begin{tabular}{lccccc}
\toprule
\textbf{Model} & \textbf{UCPI} & \textbf{UIOA} & \textbf{PASV} & \textbf{PSSA} & \textbf{CPV} \\
\midrule
\multicolumn{5}{l}{\textit{Small Language Models}}\\
\midrule
BERT Base         & 0.562 & 0.484 & 0.500 & \textbf{0.614} & 0.293 \\
BERT Large        & 0.528 & 0.437 & 0.543 & 0.529 & 0.306 \\
DeBERTa V3 Base   & 0.675 & 0.505 & 0.580 & 0.523 & 0.400 \\
DeBERTa V3 Large  & \textbf{0.696} &\textbf{0.649} & \textbf{0.683} & 0.547 & 0.460 \\
RoBERTa Base      & 0.693 & 0.547 & 0.674 & 0.515 & \textbf{0.486} \\
RoBERTa Large     & 0.682 & 0.630  & 0.680 & 0.505 & 0.444 \\
DistilBERT Base   & 0.599 & 0.547 & 0.564 & 0.450 & 0.400 \\
\midrule
\multicolumn{5}{l}{\textit{Large Language Models}}\\
\midrule
GPT 4o Mini                         & 0.543 & 0.547 & 0.632 & 0.458 & 0.324 \\
GPT 4.1 Mini                        & \textbf{0.702} & 0.469 & 0.717 & \textbf{0.479} & 0.371 \\
Gemini 2.0 Flash                    & 0.639 & \textbf{0.604} & 0.722 & 0.452 & 0.444 \\
Llama 3.3 70B                      & 0.569 & 0.427 & \textbf{0.738} & 0.415 & \textbf{0.496} \\
Gemma 3 27B it & 0.682 & 0.395 & 0.667 & 0.424 & 0.407 \\
\midrule
\multicolumn{5}{l}{\textit{Baselines}}\\
\midrule
Random   & 0.279 & 0.205 & 0.122 & 0.179 & 0.162 \\
LR with BoW   & 0.581 & 0.477 & 0.595 & 0.424 & 0.376 \\
\bottomrule
\end{tabular}
\caption{LMs’ F\textsubscript{1} scores on binary intent classification across five categories, compared to random and BoW logistic regression baselines.}
\label{tab:intent-per-class-slms-results}
\end{table}

\paragraph{Multilabel Detection.}

We show results for this task in Table~\ref{tab:intent-multi-slms-results}. DeBERTa V3 and RoBERTa continued to perform best among the SLMs, achieving the highest weighted F\textsubscript{1} scores. The best-performing LLM, LlaMA 3.3 70B, lagged noticeably behind the top SLMs. Surprisingly, the logistic regression baseline (using a one-vs-rest strategy) outperformed most LLMs. However, fine-tuned SLMs demonstrated superior ability to capture the complexity of co-occurring intent labels, underscoring the effectiveness of supervision.

\begin{table}[h!]
\centering
\scriptsize
\begin{tabular}{lcc}
\toprule
\textbf{Model} & \textbf{Micro F\textsubscript{1}}  & \textbf{Weighted F\textsubscript{1}} \\
\midrule
\multicolumn{2}{l}{\textit{Small Language Models}}\\
\midrule
BERT Base        & 0.421  & 0.414  \\
BERT Large        & 0.578  & 0.521  \\
DeBERTa V3 Base   & 0.812  & 0.804 \\
DeBERTa V3 Large  & \textbf{0.817} & 0.815  \\
RoBERTa Base      & 0.813 & \textbf{0.821}  \\
RoBERTa Large     & 0.775  & 0.808  \\
DistilBERT Base   & 0.759 & 0.769  \\
\midrule
\multicolumn{2}{l}{\textit{Large Language Models}}\\
\midrule
GPT 4o Mini              & 0.446 & 0.457 \\
GPT 4.1 Mini             & 0.489 & 0.498 \\
Gemini 2.0 Flash           & 0.410 & 0.404 \\
Llama 3.3 70B            & \textbf{0.542} &\textbf{0.570} \\
Gemma 3 27B it              & 0.440 & 0.485 \\
\midrule
\multicolumn{2}{l}{\textit{Baselines}}\\
\midrule
Random     & 0.192 & 0.201 \\
LR with BoW (OvR)   & 0.503 & 0.491 \\
\bottomrule
\end{tabular}
\caption{Performance of LMs on multilabel intent classification compared to baselines: a random classifier and a one-vs-rest logistic regression using BoW approach.}
\label{tab:intent-multi-slms-results}
\end{table}

\section{Intent-Augmented Disinformation Detection}

Inoculation theory, introduced by \citet{mcguire1964inducing}, uses a biological metaphor. It suggests that, just as people can be protected against viruses through vaccines, they can also be “vaccinated” to resist persuasive messages
\cite{mcguire1964inducing}. An inoculation message has two parts: a \textit{threat} and \textit{refutational preemption}. The threat alerts individuals that a persuasive attack is coming \cite{lewandowsky2017beyond}. Refutational preemption (or prebunking) involves providing people with arguments or tools to resist persuasive attacks, helping them better recognize and respond to such attempts \cite{pfau2005inoculation}. Building on this theory and its applicability to improve disinformation detection \cite{traberg2022psychological}, we pose the following research question: \textit{Does inoculating LLMs against malicious intent improve their disinformation detection performance in a zero-shot setting?}

To answer this question, we designed an intent-based inoculation (an IBI, we call it also intent-augmented reasoning) experiment in which the threat is an information that the text might hide malicious intent. Refutational preemption in the IBI consists of the LLM-generated analysis of intent. This analysis is generated by utilizing knowledge about types of malicious intent from our taxonomy.


\subsection{Datasets}
We rigorously evaluate intent-augmented reasoning on the MALINT dataset and 5 additional datasets covering diverse topics, text genres and languages:
\begin{itemize}[nosep, leftmargin=*]
\item \textbf{ISOT FakeNews}: Thousands of real/fake news articles from reputable sources and sites flagged by PolitiFact\footnote{PolitiFact is a nonprofit fact-checking organization.} \cite{ahmed2018detecting, ahmed2017detection}.
\item \textbf{CoAID}: COVID-19 misinformation dataset with news and social media posts \cite{cui2020coaid}.
\item \textbf{EUDisinfo}: The latest English disinformation dataset collected from the EUvsDisinfo database\footnote{The EUvsDisinfo comprises 19,455 disinformation cases (number as of October 2, 2025). Link: \url{https://EUvsDisinfo.eu/disinformation-cases/}} \cite{modzelewski2025pcot}.
\item \textbf{ECTF}: COVID-19 fake post detection dataset from Platform X (Twitter) \cite{bansal2021combining}.
\item \textbf{EUvsDisinfo}: Multilingual EUvsDisinfo's texts with pro-Kremlin propaganda \cite{leite2024euvsdisinfo}.
\end{itemize}

We used five datasets (including MALINT) to assess the usefulness of intent-based reasoning across genres and temporal splits in English. To evaluate its cross-lingual generalizability, we also used the EUvsDisinfo texts, splitting it into six language-specific datasets: German, French, Polish, Estonian, Russian, and Spanish.

\subsection{Intent-based Inoculation Design}
\label{subsec:ibi_design}

As a first step in the IBI framework, the model  \( M \) generates a structured intent analysis of the input text \( T \). This is a multilabel task over a predefined taxonomy of malicious intents \( I = \{i_1, i_2, \ldots, i_m\} \), accompanied by natural language explanations. To facilitate this, the model receives the text \( T \), external knowledge \( K_I \) describing common types of malicious intent, and \( G_A \) that provides task guidance and desired output structure. We define the intent analysis prompt as:
\begin{equation}
X_T = (T, K_I, G_A)
\end{equation}
The model \( M \) produces a structured output:
\begin{equation}
A_I(T) = \{i_j : (r_{i_j}, R_{i_j}) \mid i_j \in I\},
\end{equation}
where each \( r_{i_j} \in \{\texttt{Yes}, \texttt{No}\} \) is a binary label indicating whether intent \( i_j \) is present in the text, and \( R_{i_j} \) is the accompanying rationale.

Formally, we define:
\begin{equation}
A_I(T) \sim M(X_T) = M(T, K_I, G_A).
\end{equation}


To test if intent-inoculated LLMs improve disinformation detection, we design an inoculation prompt with a threat and a refutational preemption:
\begin{itemize}[nosep, leftmargin=*]
  \item The \textbf{threat} \( \theta \) is a textual warning that the input text may contain hidden malicious intent.
  \item The \textbf{refutational preemption} is constructed from the previously generated analysis \( A_I(T) \).
\end{itemize} These elements are combined with the original text \( T \), and detection-specific task guidelines \( G_I \). The full IBI input is then:
\begin{equation}
Z_T = (T, \theta, A_I(T), G_I).
\end{equation}
The model \( M \) uses this input for binary detection, indicating whether \( T \) is considered disinformative:
\begin{equation}
\hat{y}_T \sim M(Z_T) = M(T, \theta, A_I(T), G_I).
\end{equation}

This design of experiments allows us to answer our research question. By integrating the threat component and a LLM-generated refutational preemption into the prompt, the IBI leverages ideas built upon inoculation theory to prepare the model for detecting disinformative content.


\subsection{Experimental Setup}

We created five test sets by randomly sampling about 400-500 texts from each of the datasets. Moreover, we sampled approximately 3,000 texts from EUvsDisinfo, creating test sets of about 500 per language across six languages. Table \ref{tab:test_dataset} reports class proportions across all test sets.

\begin{table}[h]
\centering
\scriptsize
\begin{tabular}{lcc}
\toprule
\textbf{Dataset} & \textbf{Disinformation} & \textbf{Credible} \\
\midrule
MALINT & 30\% & 70\% \\
ISOT Fake News & 55\% & 45\% \\
CoAID & 21\% & 79\%  \\
EUDisinfo & 33\% & 67\%  \\
ECTF & 41\% & 59\%  \\
EUvsDisinfo & 49\% & 51\%  \\
\bottomrule
\end{tabular}
\caption{Class distribution across test datasets. The percentages represent the original distribution of disinformation and credible content within each dataset.}
\label{tab:test_dataset}
\end{table}

Our experiments were conducted on the same five LLMs that we used for experiments in \mbox{section \ref{sec:lms_evaluation}}. In these experiments, we again set the temperature hyperparameter to zero for all models. We focused on zero-shot settings with LLMs to evaluate the effectiveness of the IBI when texts lack explicit information about malicious intent. Additionally, prior studies show that LLMs can outperform supervised models such as BERT in disinformation detection \cite{pelrine2023towards, bang2023multitask}. \citet{lucas2023fighting} also found that fine-tuned BERT models perform worse on unseen data compared to zero-shot with LLMs, which was later confirmed by \citet{modzelewski2025pcot}.

To thoroughly evaluate IBI, we used two data splits on English data: (a) a genre-based split, separating long-form news articles from social media posts (from Platform X, formerly Twitter), and (b) a temporal split, comparing texts published before and after the LLMs’ knowledge cutoffs. The temporal split is possible because EUDisinfo and MALINT include post-cutoff articles not seen during model training. Following \citet{lucas2023fighting}, this setup enables a rigorous evaluation of IBI across two genres and, crucially, on unseen data, providing a more realistic test of its generalization. IBI was further evaluated on six languages to highlight its cross-lingual generalization.

In our study, we compare IBI to three competitive methods that were best on human-annotated datasets from study by \citet{lucas2023fighting}. Below short description of chosen methods:
\begin{itemize}[nosep, leftmargin=*]
\item \textit{VaN} – A minimal baseline prompt with direct instructions to the LLM \cite{lucas2023fighting}.
\item \textit{Z-CoT} – Extends \textit{VaN} with zero-shot chain-of-thought reasoning \cite{kojima2022large}.
\item \textit{DeF-SpeC} – Emphasizes contextual, deductive, and abductive reasoning \cite{lucas2023fighting}, improving multi-step reasoning \cite{bang2023multitask}.
\end{itemize}
To learn more about these methods, prompts, and how we adapted them to our IBI, see Appendix \ref{appendix:lucas_methods}. 

We use F\textsubscript{1} for the positive class as evaluation metric. To assess significance between IBI and baselines, we used McNemar's test (see \mbox{section \ref{subsec:ibi_results_discussion}} and \mbox{Appendix~\ref{appendix:mcnemar}} for details), a standard method for comparing two models on the same binary task \cite{dietterich1998approximate, dror2018hitchhiker}, widely used in NLP \cite{blitzer2006domain, card2020little}.

\subsection{Results and Discussion}
\label{subsec:ibi_results_discussion}

\paragraph{MALINT: Dataset-Specific Results}
Table \ref{tab:ibi_results_malint} compares baseline prompting strategies (Base) with their intent-based inoculation (IBI) counterparts on the MALINT dataset. Across all models, IBI consistently improves disinformation detection, with average gains ranging from around 2\% for GPT-4o Mini up to 8\% for Gemini 2.0 Flash.

We analyzed the effect of intent prediction correctness on disinformation detection for each intent type separately. For each model and intent type, we computed F\textsubscript{1} scores separately for instances where the intent was predicted correctly versus incorrectly. This allowed us to measure the potential impact of separate intent knowledge on the model’s ability to detect disinformation with IBI. The results show that correct intent predictions generally improve F\textsubscript{1} scores, indicating that accurate intent understanding boosts disinformation detection with IBI. Some exceptions exist, such as UIOA for certain models, where higher F\textsubscript{1} scores with IBI occur even when intent is mispredicted, suggesting that other correctly predicted intents can partially compensate. 

Appendix \ref{sec:ablation_impact_intent_on_ibi} shows the ablation on intent’s influence on disinformation prediction, and remaining dataset-specific results are in Appendix \ref{appendix:ibi_results_specific_datasets}.

\paragraph{Genre and Temporal Split Results} 
Table~\ref{tab:ibi_f1_overall_scores} compares the baseline prompting strategies (Base) with their IBI counterparts. Across 75 evaluation scenarios (5 models × 3 prompting strategies × 5 settings: overall comparison, articles, social media posts, prior-cutoff, and post-cutoff), IBI leads to improved performance in approximately 90\% of cases. On average, the overall performance increases by 9\%. McNemar’s test indicates that, in nearly all scenarios, IBI significantly outperforms the baselines on the overall dataset at the 0.01 significance level. For Llama 3.3 70B, the difference is still significant, though at the 0.05 level.

The greatest gains are observed in longer-form articles. We hypothesize that the extended context in articles offers language models more opportunity to identify and reason about malicious intent.

Notably, IBI improves performance not only on data that may have been present during LLM pretraining, but also on unseen content published after the models' knowledge cutoffs. While performance gains are evident in both pre- and post-cutoff subsets, LLMs show greater difficulty with the latter.

Overall, these findings support our central hypothesis: incorporating intent-based inoculation enhances zero-shot disinformation detection. Improvements are consistent across data types, temporal settings, and prompting strategies.


\paragraph{Multilingual Evaluation}

Table~\ref{tab:ibi_vs_base_per_language} shows the averaged F\textsubscript{1} scores across all models and methods for each language on the EUvsDisinfo dataset. IBI consistently improves performance over the Base setup for all six languages, with the largest gains observed in Estonian. These results indicate that intent-augmented reasoning enhances model disinformation detection capabilities across languages. Overall, IBI demonstrates a clear advantage in cross-lingual disinformation detection, achieving on average a 20\% improvement over baseline methods. More detailed results in Appendix \ref{appendix:cross-lingual-res-ibi}.

\begin{table}[h!]
\centering
\scriptsize
\renewcommand{\arraystretch}{1.1}
\begin{tabular}{lcc}
\hline
\textbf{Language} & \textbf{Base} & \textbf{IBI} \\
\hline
German   & 0.794 & \perc{0.794}{0.911} \\
Spanish  & 0.683 & \perc{0.683}{0.828} \\
Estonian & 0.716 & \perc{0.716}{0.892} \\
French   & 0.611 & \perc{0.611}{0.749} \\
Polish   & 0.709 & \perc{0.709}{0.846} \\
Russian  & 0.619 & \perc{0.619}{0.735} \\
\hline
\end{tabular}
\caption{Comparison of Base vs IBI performance across languages. Results present averaged F\textsubscript{1} scores  over all models and methods. See Appendix~\ref{appendix:cross-lingual-res-ibi} for the table with standard deviations.}
\label{tab:ibi_vs_base_per_language}
\end{table}

\begin{table*}[!ht]
\scriptsize
\centering
\begin{tabular}{lcccccccccc}
\toprule
 & \multicolumn{2}{c}{\textit{GPT 4o Mini}} 
 & \multicolumn{2}{c}{\textit{GPT 4.1 Mini}} 
 & \multicolumn{2}{c}{\textit{Gemini 2.0 Flash}} 
 & \multicolumn{2}{c}{\textit{Gemma 3 27b it}} 
 & \multicolumn{2}{c}{\textit{Llama 3.3 70B}} \\
\cmidrule(lr){2-3} \cmidrule(lr){4-5} \cmidrule(lr){6-7} \cmidrule(lr){8-9} \cmidrule(lr){10-11}
 & Base & IBI & Base & IBI & Base & IBI & Base & IBI & Base & IBI \\
\midrule
VaN       & 0.815 & \perc{0.815}{0.856} 0.856 &  0.825 & \perc{0.825}{0.873}  & 0.789 & \perc{0.789}{0.855}  & 0.783 & \perc{0.783}{0.820}  & 0.836 & \perc{0.836}{0.863}  \\
Z-CoT     & 0.836 & \perc{0.836}{0.849}   & 0.810 & \perc{0.810}{0.861}   & 0.751 & \perc{0.751}{0.837} & 0.782 & \perc{0.782 }{0.806}  & 0.807 & \perc{0.807}{0.865}   \\
DeF\_Spec & 0.887 & \perc{0.887}{0.877}  & 0.870 & \perc{0.870}{0.879}  & 0.843 & \perc{0.843}{0.881}  & 0.812 & \perc{0.812}{0.846}  & 0.806 & \perc{0.806}{0.871 }  \\
\bottomrule
\end{tabular}
\caption{F\textsubscript{1} scores on MALINT for competitive prompting methods and their improvement with IBI.}
\label{tab:ibi_results_malint}
\end{table*}

\begin{table*}[!ht]
\scriptsize
\centering
\begin{tabular}{lcccccccccc}
\toprule
 & \multicolumn{2}{c}{\textbf{Overall}} & \multicolumn{2}{c}{\textbf{Articles}} & \multicolumn{2}{c}{\textbf{Posts}} & \multicolumn{2}{c}{\textbf{Prior Cutoff}} & \multicolumn{2}{c}{\textbf{Post Cutoff}} \\
\cmidrule(lr){2-11}
 & Base & IBI & Base & IBI & Base & IBI & Base & IBI & Base & IBI \\
\midrule
\multicolumn{11}{l}{\textit{GPT 4o Mini}}\\

\quad VaN & 0.736 & \perc{0.736}{0.828} & 0.754 & \perc{0.754}{0.862}  & 0.703 & \perc{0.703}{0.755}  & 0.727 & \perc{0.727}{0.821}  & 0.762 & \perc{0.762}{0.846}  \\

\quad Z-CoT & 0.740 & \perc{0.740}{0.826} & 0.764 & \perc{0.764}{0.854}  & 0.692 & \perc{0.692}{0.766}  & 0.724 & \perc{0.724}{0.823}  & 0.786 & \perc{0.786}{0.833} \\

\quad DeF-SpeC & 0.746 & \perc{0.746}{0.792} & 0.782 & \perc{0.782}{0.817}  & 0.682 & \perc{0.682}{0.742}  & 0.712 & \perc{0.712}{0.771}  & 0.843 & \perc{0.843}{0.850} \\

\midrule
\multicolumn{11}{l}{\textit{GPT 4.1 Mini}}\\

\quad VaN & 0.698 & \perc{0.698}{0.751} & 0.718 & \perc{0.718}{0.772}  & 0.659 & \perc{0.659}{0.705}  & 0.672 & \perc{0.672}{0.709}  & 0.767 & \perc{0.767}{0.862}  \\

\quad Z-CoT & 0.673 & \perc{0.673}{0.748} & 0.685 & \perc{0.685}{0.765}  & 0.649 & \perc{0.649}{0.712}  & 0.640 & \perc{0.640}{0.710}  & 0.757 & \perc{0.757}{0.849} \\

\quad DeF-SpeC & 0.748 & \perc{0.748}{0.780} & 0.780 & \perc{0.780}{0.803}  & 0.686 & \perc{0.686}{0.732}  & 0.720 & \perc{0.720}{0.752}  & 0.828 & \perc{0.828}{0.856} \\

\midrule
\multicolumn{11}{l}{\textit{Gemini 2.0 Flash}}\\

\quad VaN & 0.701 & \perc{0.701}{0.762} & 0.703 & \perc{0.703}{0.803}  & 0.699 & \perc{0.699}{0.677}  & 0.682 & \perc{0.682}{0.731}  & 0.754 & \perc{0.754}{0.851}  \\

\quad Z-CoT & 0.670 & \perc{0.670}{0.733} & 0.667 & \perc{0.667}{0.763}  & 0.675 & \perc{0.675}{0.670}  & 0.646 & \perc{0.646}{0.694}  & 0.736 & \perc{0.736}{0.838} \\

\quad DeF-SpeC & 0.767 & \perc{0.767}{0.803} & 0.795 & \perc{0.795}{0.835}  & 0.710 & \perc{0.710}{0.738}  & 0.749 & \perc{0.749}{0.787}  & 0.814 & \perc{0.814}{0.847} \\
\midrule

\multicolumn{11}{l}{\textit{Gemma 3 27b it}}\\

\quad VaN & 0.694 & \perc{0.694}{0.773} & 0.684 & \perc{0.684}{0.801}  & 0.711 & \perc{0.711}{0.710}  & 0.662 & \perc{0.662}{0.750}  & 0.782 & \perc{0.782}{0.830}  \\

\quad Z-CoT & 0.622 & \perc{0.622}{0.767} & 0.671 & \perc{0.671}{0.793}  & 0.516 & \perc{0.516}{0.711}  & 0.561 & \perc{0.561}{0.746}  & 0.775 & \perc{0.775}{0.822} \\

\quad DeF-SpeC & 0.739 & \perc{0.739}{0.791} & 0.742 & \perc{0.742}{0.825}  & 0.734 & \perc{0.734}{0.720}  & 0.712 & \perc{0.712}{0.769}  & 0.815 & \perc{0.815}{0.851} \\

\midrule
\multicolumn{11}{l}{\textit{Llama 3.3 70B}}\\

\quad VaN & 0.756 & \perc{0.756}{0.770} & 0.762 & \perc{0.762}{0.796}  & 0.744 & \perc{0.744}{0.717}  & 0.730 & \perc{0.730}{0.738}  & 0.824 & \perc{0.824}{0.856}  \\

\quad Z-CoT & 0.736 & \perc{0.736}{0.781} & 0.739 & \perc{0.739}{0.804}  & 0.730 & \perc{0.730}{0.733}  & 0.714 & \perc{0.714}{0.748}  & 0.793 & \perc{0.793}{0.867} \\

\quad DeF-SpeC & 0.716 & \perc{0.716}{0.762} & 0.723 & \perc{0.723}{0.788}  & 0.702 & \perc{0.702}{0.707}  & 0.684 & \perc{0.684}{0.720}  & 0.798 & \perc{0.798}{0.872} \\

\midrule

\textbf{Average} & 0.716 & \perc{0.716}{0.778} & 0.731 & \perc{0.731}{0.805} & 0.686 & \perc{0.686}{0.720} & 0.689 & \perc{0.689}{0.751} & 0.789 & \perc{0.789}{0.849} \\

\bottomrule
\end{tabular}
\caption{Results with F\textsubscript{1} scores  for five LLMs. The \textit{Base} columns shows the competitive method results, while the \textit{IBI} columns presents results for prompts adapted to the Intent-based Inoculation. 
}
\label{tab:ibi_f1_overall_scores}
\end{table*}

\section{Related Work}

\paragraph{Intent.}
Intent (or \textit{intention}) discovery is multifaceted problem addressed from different perspectives. It can be purely textual ~\citep{Xu2023} or multimodal, where text is analyzed alongside images~\citep{kruk-etal-2019-integrating} or videos~\citep{maharana-etal-2022-multimodal}. 
Some studies focus on uncovering the relationship between intention and behavior~\citep{Conner2022-CONUTI-2} or how intentions guide people to achieve a goal~\citep{suchodoletz2011}, while others aim to identify and categorize intentions.
~\citet{10.1093/ct/qtac021} defines a conceptualization that connects actors, intentions, and techniques for creating and disseminating disinformation content. They 
identify four literature-based categories of intentions: \textit{delegitimization}, \textit{mobilization}, \textit{ideological motivations}, and \textit{financial gain}. They present a framework that can be used to conceptualize disinformation, but they do not annotate a dataset nor provide a baseline for intention detection and any experiments with LMs. \citet{10.1145/3627673.3679799} use public datasets of real/fake articles and annotate them using a T5-based model with agent intent classes: \textit{Public}, \textit{Emotion}, \textit{Individual}, \textit{Popularize}, \textit{Clout}, \textit{Conflict}, \textit{Smear}, \textit{Bias}, \textit{Connect}. 
Moreover, they show that incorporating intent features improves misinformation detection performance in their T5-based framework. 

~\citet{gupta2021truthbotautomatedconversationaltool} explore user intent behind a query and proposes a fact-checking chatbot to counter the spread of fake news, but does not investigate whether intent could be leveraged to improve disinformation detection.
Instead, ~\citet{Zhou_2022} focuses on the intent of spreading fake news. After labelling a dataset algorithmically, they evaluate the effectiveness of employing user propagation intent to detect fake news using Heterogeneous Graph Neural Networks, achieving a slight improvement concerning the state-of-the-art. They also found that most fake news spreaders don't do it intentionally, so focusing on disinformation agents' intent could be more meaningful.
 

\paragraph{Disinformation.}



Prior work on text-based disinformation detection has relied mainly on supervised classification of false or misleading content. Early datasets include LIAR \cite{wang2017liar}, with short statements labeled for veracity, and FakeNewsNet \cite{shu2020fakenewsnet}, which links news with social and temporal context. Most systems model disinformation as binary classification (real vs. fake), using fine-tuned models like BERT \cite{khan2021benchmark}. Surveys highlight that fake news is intentionally misleading \cite{shu2017fake}. As a result, RMDM \cite{nguyen2023rmdm}, a Vietnamese dataset, adds four labels (real, misinfo, disinfo, malinfo) to separate unintentional errors from deliberate harm. The only dataset annotated with intent and disinformation was introduced by \citet{modzelewski2024mipd}. However, the dataset is limited to Polish, and their study reports only simple baseline results for intent detection in that language. 
Moreover, recent work shows that pretraining on related tasks (e.g., fine-grained sentiment) can enhance disinformation detection \cite{pan2024enhancing}.

To the best of our knowledge, MALINT is the first human-annotated English dataset to label disinformation and malicious intents of disinformation agents (see Appendix \ref{appendix:table-comparison-benchmarks} for full comparison to other datasets). We are the first to evaluate intent classification with SLMs and LLMs, establishing baselines for intent detection in English. Our study also demonstrates, for the first time, the utility of intent-augmented reasoning with different LLMs for zero-shot disinformation detection.

\section{Conclusions}

In this study, we present \textbf{MALINT}, the first human-annotated English-language dataset that includes disinformation and malicious intent annotations. MALINT is a high-quality dataset created with accredited disinformation and fact-checking experts.

Our research provides the first systematic evaluation of malicious intent identification, comparing seven fine-tuned SLMs (e.g., BERT) and zero-shot with LLMs across binary and multilabel settings. In the multilabel setup, SLMs outperform LLMs, reaching a weighted F\textsubscript{1} of 82.1\%. In contrast, binary classification reveals that LLMs outperform fine-tuned SLMs on three intent categories.

Inspired by inoculation theory from psychology and communication studies, we designed an experiment on intent-based inoculation with LLMs. Our results show that exposure to knowledge about malicious intent significantly enhances disinformation detection performance in a zero-shot setting. IBI improves performance by 9\% on average across five English datasets and LLMs, and enhances disinformation detection in other six languages.


\section*{Limitations}

\paragraph{Datasets and Annotation.}
The MALINT dataset features five malicious intent categories, in addition to a binary classification for disinformation. While this framework provides a comprehensive representation, we acknowledge that the dataset may not exhaustively cover all possible intents. While this taxonomy offers a broad and representative framework, we acknowledge that it may not capture the full spectrum of possible intent types. Our categorization draws on prior research \cite{modzelewski2024mipd} as well as official reports from an independent non-profit organisations, like EU DisinfoLab (e.g. \citet{sessa2023connecting}). Furthermore, the taxonomy was developed in close collaboration with experts in fact-checking and debunking, ensuring informed coverage of malicious intent types that are especially relevant to the current global information landscape. 

\paragraph{Biases.}
Human annotation inherently involves some degree of subjective interpretation. In order to mitigate this issue, annotation was conducted under the supervision of professional fact-checkers. Annotators underwent extensive training, following the detailed annotation guidelines developed with experts. Each article was independently annotated by each annotator and subsequently reviewed by a supervisor. When clarification was required, the supervisor consulted the original annotator to ensure consistency and accuracy. For what concerns the other datasets employed in this study, they inevitably retain any biases present in their original annotation process.

\paragraph{Experiments.} The intent-based reasoning experiments conducted in this study rely on the fixed taxonomy of malicious intents adopted in the proposed dataset. Although this specific framework may contain inherent bias, our categorization was rigorously validated by professional fact-checkers. It builds on established research \cite{modzelewski2024mipd} and incorporates insights from independent non-profit organizations such as EU DisinfoLab (e.g. \citet{sessa2023connecting}), ensuring both empirical grounding and practical relevance.

Although we worked with a limited set of models, we deem this selection adequately diverse. It includes both Small and Large Language Models, incorporating open-source and closed models and covering multiple providers such as OpenAI, Google and Meta.

\section*{Ethics}
\paragraph{Dataset and Annotation.} The MALINT dataset comprises data extracted from publicly accessible online sources and is free from copyright restrictions. It does not contain personally identifiable information and is intended solely for research applications. The dataset will be released under the CC BY 4.0 license.

No crowdsourcing was used at any stage of the annotation process. All annotators were hired by affiliated institutions and received appropriate compensation. The annotation process was designed to ensure fair and unbiased work practices, with expert oversight throughout and approval obtained from the relevant ethics board. We also made efforts to maintain gender balance among annotators to promote diverse perspectives.

Our research aims to support society by helping fact-checkers, researchers, and public-interest efforts. However, the dataset and insights we provide could also be misused. For example, malicious actors might study the MALINT dataset to understand detection patterns and create more convincing or harder-to-spot disinformation. We strongly encourage ethical use. Any system built using our data should include transparency, oversight, and safeguards to reduce the risk of abuse.

\paragraph{Computational resources.} The deployment of large language models raises environmental concerns, due to significant computational requirements. Our methodology mitigates these issues by focusing on model inference rather than training from scratch, thereby substantially reducing computational demands. Our experiments primarily relied on third-party API services, with computational resources managed by external providers while fine-tuning was conducted only on small language models. All computing resources were provided by university and reserved exclusively for research purposes.

\section*{Acknowledgements}

This research was supported by the projects: Infotester4Education (full title: Development and implementation of AI education methods and digital tools supporting tackling disinformation, number:  2023-2-PL01-KA220-HED-000180856) within the framework of Cooperation Partnership for Higher Education, ERASMUS+, and EUonAIR project (number 101177370, ERASMUS-EDU-2024-EUR-UNIV-1), within the framework of the EUonAIR Centre of Excellence in Responsible AI in Education, co-funded by the European Commission.

Giovanni Da San Martino would like to thank the Qatar National Research Fund, part of Qatar Research Development and Innovation Council (QRDI), for funding this work  by grant NPRP14C0916-210015. 
He also would like to thank the European Union under the National Recovery and Resilience Plan (NRRP), Mission 4 Component 2 Investment 1.3 - Call for tender No. 341 of March 15, 2022 of Italian Ministry of University and Research – NextGenerationEU; Code PE00000014, Concession Decree No. 1556 of October 11, 2022 CUP D43C22003050001, Progetto ``SEcurity and RIghts in the CyberSpace (SERICS)'' - Spoke 2 Misinformation and Fakes - DEcision supporT systEm foR cybeR intelligENCE (Deterrence) for also funding this work.


\appendix

\section{Annotatin Guidelines}
\label{appendix:guidelines}

Our methodology and annotation guidelines were designed to standardize the assessment of articles for disinformation content, aiming to reduce subjectivity and enable comprehensive analysis. Utilizing these annotation guidelines, we analyzed numerous articles to identify disinformation. The methodology was developed in cooperation with analysts (fact-checking and debunking experts) employed in the project based on their experience as experts, scientific knowledge available on the subject, and the experience of other institutions and organizations involved in research and detection of disinformation. The methodology improved throughout the project and subsequent testing to best reflect the disinformation environment. All authors of this methodology have at least three years of experience working for fact-checking or debunking organizations accredited by the International Fact-Checking Network. Moreover, our methodology and annotation guidelines draw on similar work on the annotation of disinformation, such as the guidelines presented by \citet{modzelewski2024mipd}.

\paragraph{Main Assumptions of the Methodology.}
Creating a uniform methodology and guidelines aims to guarantee the quality of the assessments made by annotators and minimize their subjectivity. 

The analysis of articles is carried out mainly via the debunking technique, with the auxiliary use of the fact-checking technique. These terms for this methodology are defined in a manner analogous to the methodology developed for the NATO Strategic Communication Centre of Excellence \cite{pamment2021fact}. Fact-checking is the long-standing process of checking that all facts in a piece of writing, news article, or speech are correct. Debunking refers to exposing falseness or manipulating systematically and strategically (based on a chosen topic, classifications of selected techniques, narrative). 

\paragraph{Preparation of Articles for Evaluation.} The first step is to select web portals from which articles on particular topics will be taken. Among them are both mainstream media and those presenting the alternative current. This is to ensure access to enough reliable as well as unreliable content. Each portal will be assigned to one of three categories, determining its credibility. This will be done by a team of experts by consensus. Assessing the credibility of a website requires an in-depth analysis of the content posted on it regularly, as well as checking it in reliable sources, including via the Media Bias/Fact Check search engine. The source's rating will not be visible to annotators. The analysis consists in selecting the category that best suits a given domain:
\begin{itemize}[nosep, leftmargin=*]
    \item {\textbf{Reliable}} - sources that are reliable/publishing reliable content on a specific topic, in particular traditional news portals.
    \item {\textbf{Unreliable}} - sources publishing unreliable content, typically disinformation, e.g., all domains financed by the Kremlin, sites containing conspiracy theories, etc.
    \item {\textbf{Mixed/Biased}} - partially or potentially biased websites that may present false information on specific issues, e.g., typically political websites, and blog collections.
\end{itemize}

\paragraph{Thematic Category.}

Before the analysis begins, articles will be assigned to eight topics. This will be done manually with the help of keywords through searches on selected web portals. Thematic categories were pre-defined. The selection of topics was based on EU DisinfoLab's cross-cutting report on disinformation in Europe \cite{sessa2023connecting}. It is based on expert studies from 20 countries. 

\begin{itemize}[nosep, leftmargin=*]
    \item Anti-Europeanism and anti-Atlanticism (anti-EU, anti-NATO)
    \item Anti-migration and xenophobia
    \item Climate change and the energy crisis 
    \item Health (including COVID-19 and vaccines)
    \item Institutional and media distrust (public institutions)
    \item Gender-based disinformation
    \item Ukraine war and refugees 
    \item Disinformation about LGBTQIA+
\end{itemize}

\paragraph{Content Analysis.} 

The next step requires analyzing the entire article's content and recognizing whether the information is accurate or disinformative. If the article provides only factual information, it is marked as “credible information.” Selecting this category ends the assessment of the article. When information in the article is unreliable and misleads the recipients, content is considered disinformative. The unintentional dissemination of false information is known as misinformation. However, even unintentional dissemination of false information without the goal of manipulating recipients can fuel disinformation. Disinformation is particularly difficult to detect as the author’s intention is usually unspecified, and in most cases, it can only be presumed. Therefore, for this study, we assume that any form of false or manipulative information is considered disinformation.

For these guidelines, the definition of disinformation provided by the European Commission High-Level Group of Experts on False News and Disinformation on the Internet (HELG) will be used, as it covers all four aspects and does not exclude potentially harmful content presented in the form of political advertising or satire, as presented in the EU Code of Practice. The definition is as follows \cite{de2018multi}: 
\begin{quote}
    \itshape 
    `` All forms of false, inaccurate, or misleading information designed, presented, and promoted to intentionally cause public harm or for profit.''
    \hfill  
\end{quote}

However, a necessary supplement to this definition is taking into account the European Union Code of Practice on Disinformation, according to which disinformation is defined as: "verifiable false or misleading information which, cumulatively, (a) is created, presented and disseminated for economic gain or to intentionally deceive the public; and (b) may cause public harm, intended as threats to democratic political and policymaking processes as well as public goods such as the protection of EU citizens' health, the environment, or security". \cite{eu2022strengthdiscode}. The detected information must be verifiable, which means that it can be proved untrue, and, therefore, it cannot be, for example, a yet unproven theory or opinion, as long as it is not intended to mislead the recipients. In summary, disinformation is intentionally misleading by providing misleading or false information \cite{eu2020communication}. Unlike disinformation, misinformation is \textit{misleading information shared by people who do not recognize it as such} \cite{de2018multi}. However, as noted earlier, misinformation and disinformation are treated as a single category under "disinformation."

When a given content is not verifiable (reliable/disinformative), it is marked as the "Hard to say" category. Indicating this category ends the assessment. 
Below, we present the main categories:

\begin{itemize}[nosep, leftmargin=*]
    \item Credible information
    \item Disinformation 
    \item Hard to say
\end{itemize}

\paragraph{Annotation of Malicious Intent.} The study of the malicious intentions of the disinformation content creators is potentially the most subjective element of the analysis, and therefore it is particularly important to develop precise components of the assessment. This allows for maintaining uniformity of the analysis carried out by different annotators.

In this methodology, understanding the intention behind disinformation is crucial for effectively analyzing it. Disinformation, according to our definition, is always spread intentionally, emphasizing the significance of comprehending the motives driving its dissemination. It encapsulates the broader goal of the author, which guides the specific narratives they employ to achieve that goal. Authors of disinformation have some purpose in creating it. It is in this category that we try to answer the question: what is the purpose of spreading disinformation by a particular author? The task type is defined as an exhaustive list with multiple choice options (multilabel). Below are possible choices:

\begin{itemize}[nosep, leftmargin=*]
    \item \textbf{Undermining the credibility of public institutions} - The goal of many disinformation authors is to destroy trust in public institutions. This can be done by undermining official communications, insinuating bad intentions or falsely exposing corruption (e.g., accusing governments of population control with vaccines). The idea is to make citizens disbelieve in the effectiveness of their own state, undermine the sense of its existence or actively fight against it. This is ultimately meant to lead to resentment of the system, thus undermining the very essence of Western democracies. As a result, it becomes easier to spread false information, and the public's resistance to outside influence decreases.
    \item \textbf{Changing political views} - Influencing voter preferences is a common procedure used by disinformation authors. Changing political beliefs is aimed at strengthening one side of a political dispute and arousing resentment against the others. It usually involves the simultaneous promotion of politicians from extremist movements, which are treated as an alternative to the major parties. It is often based on the portrayal of mainstream politicians as corrupt and evil to the bone (e.g., portraying them as traitors to the nation, dependent on the outside influence of global elites).
    \item \textbf{Undermining international organizations and alliances} - Undermining the credibility of international institutions is often part of disinformation activities carried out by external forces (e.g., Russia). These are aimed at breaking up alliances of democratic states to facilitate propaganda efforts by authoritarian states (e.g., portraying NATO as an aggressor that will drag peaceful states into war). Of course, numerous extreme political movements also have an interest in shattering trust in international institutions. This is part of a populist influence on society and a way to gain power. International institutions are then most often portrayed as entities that take away the sovereignty of member states (e.g., presenting the EU as an authoritarian organization that imposes its will on others).
    \item \textbf{Promoting social stereotypes/antagonisms} - Deepening social divisions is a frequent goal of disinformation efforts. A strongly divided society is less resistant to manipulation, and mutual distrust also promotes a collapse of confidence in the institution of the state and democracy. This causes internal problems to absorb most of the attention, giving room for external centers of influence to operate. This can take the form of reinforcing xenophobia (e.g., stirring up resentment against Ukrainian refugees and portraying them as dangerous). Aversion to specific social groups can also be exploited (e.g., portraying homosexuals as pedophiles).
    \item \textbf{Promoting anti-scientific views} - Science is a frequent enemy of disinformation authors. Science enhances critical thinking and is an important part of the strength of Western democracies. Presenting it as an enemy aids in undermining the system under which Western countries operate. Reinforcing anti-scientific attitudes also enables short-term financial gain (e.g., selling pseudo-medical remedies for various diseases). The fight against science can be based on a direct attack on scientists (e.g., the claim that vaccines are designed to depopulate humanity), but is also a significant element of conspiracy theories (e.g., medicine is not used to cure people, but to make money).
\end{itemize}

\paragraph{Double Evaluation and Consensus Establishment}

According to this methodology, all content must undergo a double evaluation. Articles are evaluated two times by two annotators, working independently of each other. The first is the student, and the second is the supervisor. The supervisor does not read the first performed assessment, but only evaluates the content according to the methodology, independently of the results of the first evaluation. The supervisor then compares the two performed assessments and makes the final decision on the choices made in the analysis process. Discrepancies spotted by the double-verification analyst are discussed by the team. Then, a common, consistent approach to content classification is established. When necessary, the lead annotator, an expert in fact-checking and debunking, can be consulted to discuss the evaluation. The final registered assessment is therefore a consensus based on the first and second assessment, and can include elements of both independent evaluations. The purpose of double verification is therefore not only to avoid the human errors but also to the standardization of the methodology's application.

\section{Small Languages Models and Hyperparameters for Malicious Intent Classification Tasks}
\label{appendix:exp_benchmark_setup}

This appendix provides details on the experimental setup and optimal hyperparameters used for multilabel classification of malicious intent categories. All models were trained and evaluated using the same dataset splits, with five intent categories:  \textit{Undermining the Credibility of Public Institutions} (UCPI), \textit{Changing Political Views} (CPV), \textit{Undermining International Organizations and Alliances} (UIOA), \textit{Promoting Social Stereotypes/Antagonisms} (PSSA), and \textit{Promoting Anti-scientific Views} (PASV).

\paragraph{Tokenization and Data Preprocessing}

For all models and tasks, we used the Hugging Face Transformers library to load both the tokenizer and the model. The input content was tokenized with the following settings:
\begin{itemize}[nosep, leftmargin=*]
    \item Truncation: True
    \item Padding: True
    \item Max Length: 256 tokens
\end{itemize}

\paragraph{Training and Hyperparameter Search}

The training procedure involved feeding data into a training loop using the \texttt{Trainer} API. Hyperparameter tuning was performed over the following grid:

\begin{itemize}[nosep, leftmargin=*]
    \item \textbf{Learning rate:} \{1e-5, 2e-5, 3e-5, 4e-5, 5e-5\}
    \item \textbf{Warmup ratio:} \{0.06, 0.1\}
    \item \textbf{Weight decay:} \{0.01, 0.03, 0.05, 0.1\}
\end{itemize}

\subsection{Binary Detection Per Each Class.}

All models were trained for 5 epochs with model checkpointing based on the F\textsubscript{1} score over the positive class. Other fixed hyperparameters included:
\begin{itemize}[nosep, leftmargin=*]
    \item Batch size (train/eval): 8
    \item FP16 training: Enabled
    \item Evaluation strategy: every 50 steps
    \item Save total limit: 2 checkpoints
    \item Load best model at end: True
\end{itemize}

\paragraph{Optimal Hyperparameters per Model}

The best-performing hyperparameter configuration for each model and each binary detection task is listed in Table \ref{tab:hyperparams_binary_ucpi_cpv_uioa_pssa_pasv}. Identifiers of all models given in Table \ref{tab:hyperparams_binary_ucpi_cpv_uioa_pssa_pasv} as of 21.07.2025. All models were evaluated using F\textsubscript{1} over the positive class.

 \begin{table*}[ht]
\centering
\small
\begin{tabular}{llccc}
\toprule
\textbf{Model} & \textbf{Identifier} & \textbf{Learning Rate} & \textbf{Weight Decay} & \textbf{Warmup Ratio} \\
\midrule
\multicolumn{5}{l}{\textbf{\textit{Undermining the Credibility of Public Institutions} (UCPI)}} \\
\midrule
BERT-base & google-bert/bert-base-uncased & 2e-5 & 0.03 & 0.06 \\
BERT-large & google-bert/bert-large-uncased & 1e-5 & 0.05 & 0.06 \\
RoBERTa-large & FacebookAI/roberta-large & 2e-5 & 0.05 & 0.06 \\
RoBERTa-base & FacebookAI/roberta-base & 1e-5 & 0.05 & 0.1 \\
DeBERTa-v3-large & microsoft/deberta-v3-large & 1e-5 & 0.01 & 0.06 \\
DeBERTa-v3-base & microsoft/deberta-v3-base & 2e-5 & 0.05 & 0.06 \\
DistilBERT-base & distilbert/distilbert-base-uncased & 1e-5 & 0.03 & 0.06 \\
\midrule
\multicolumn{5}{l}{\textbf{\textit{Changing Political Views} (CPV)}} \\
\midrule
BERT-base & google-bert/bert-base-uncased & 2e-5 & 0.01 & 0.1 \\
BERT-large & google-bert/bert-large-uncased & 2e-5 & 0.01 & 0.06 \\
RoBERTa-large & FacebookAI/roberta-large & 2e-5 & 0.03 & 0.1 \\
RoBERTa-base & FacebookAI/roberta-base & 2e-5 & 0.1 & 0.06 \\
DeBERTa-v3-large & microsoft/deberta-v3-large & 1e-5 & 0.05 & 0.1 \\
DeBERTa-v3-base & microsoft/deberta-v3-base & 1e-5 & 0.03 & 0.06 \\
DistilBERT-base & distilbert/distilbert-base-uncased & 2e-5 & 0.01 & 0.06 \\
\midrule
\multicolumn{5}{l}{\textbf{\textit{Undermining International Organizations and Alliances} (UIOA)}} \\
\midrule
BERT-base & google-bert/bert-base-uncased & 5e-5 & 0.03 & 0.1 \\
BERT-large & google-bert/bert-large-uncased & 1e-5 & 0.05 & 0.1 \\
RoBERTa-large & FacebookAI/roberta-large & 1e-5 & 0.05 & 0.06 \\
RoBERTa-base & FacebookAI/roberta-base & 2e-5 & 0.05 & 0.06 \\
DeBERTa-v3-large & microsoft/deberta-v3-large & 1e-5 & 0.03 & 0.06 \\
DeBERTa-v3-base & microsoft/deberta-v3-base & 3e-5 & 0.05 & 0.06 \\
DistilBERT-base & distilbert/distilbert-base-uncased & 1e-5 & 0.1 & 0.06 \\
\midrule
\multicolumn{5}{l}{\textbf{\textit{Promoting Social Stereotypes/Antagonisms} (PSSA)}} \\
\midrule
BERT-base & google-bert/bert-base-uncased & 2e-5 & 0.01 & 0.06 \\
BERT-large & google-bert/bert-large-uncased & 2e-5 & 0.03 & 0.06 \\
RoBERTa-large & FacebookAI/roberta-large & 1e-5 & 0.01 & 0.06 \\
RoBERTa-base & FacebookAI/roberta-base & 2e-5 & 0.01 & 0.06 \\
DeBERTa-v3-large & microsoft/deberta-v3-large & 1e-5 & 0.05 & 0.1 \\
DeBERTa-v3-base & microsoft/deberta-v3-base & 1e-5 & 0.03 & 0.1 \\
DistilBERT-base & distilbert/distilbert-base-uncased & 1e-5 & 0.1 & 0.1 \\
\midrule
\multicolumn{5}{l}{\textbf{\textit{Promoting Anti-scientific Views} (PASV)}} \\
\midrule
BERT-base & google-bert/bert-base-uncased & 1e-5 & 0.05 & 0.1 \\
BERT-large & google-bert/bert-large-uncased & 2e-5 & 0.01 & 0.06 \\
RoBERTa-large & FacebookAI/roberta-large & 1e-5 & 0.03 & 0.1 \\
RoBERTa-base & FacebookAI/roberta-base & 1e-5 & 0.01 & 0.1 \\
DeBERTa-v3-large & microsoft/deberta-v3-large & 1e-5 & 0.03 & 0.06 \\
DeBERTa-v3-base & microsoft/deberta-v3-base & 1e-5 & 0.05 & 0.06 \\
DistilBERT-base & distilbert/distilbert-base-uncased & 1e-5 & 0.03 & 0.1 \\
\bottomrule
\end{tabular}
\caption{Optimal hyperparameters for binary classification of all malicious intent categories.}
\label{tab:hyperparams_binary_ucpi_cpv_uioa_pssa_pasv}
\end{table*}

\subsection{Multilabel Detection.}

All models were trained for 5 epochs with model checkpointing based on the macro-weighted F\textsubscript{1} score. Other fixed hyperparameters included:
\begin{itemize}[nosep, leftmargin=*]
    \item Batch size (train/eval): 4
    \item FP16 training: Enabled
    \item Evaluation strategy: every 50 steps
    \item Save total limit: 2 checkpoints
    \item Load best model at end: True
\end{itemize}

\paragraph{Optimal Hyperparameters per Model}

The best-performing hyperparameter configuration for each model is listed in Table \ref{tab:hyperparams_multilabel}. Identifiers of all models given in Table \ref{tab:hyperparams_multilabel} as of 21.07.2025. All models were evaluated using macro-weighted F\textsubscript{1} to account for class imbalance and the multilabel nature of the task.

\begin{table*}[ht]
\centering
\small
\begin{tabular}{llccc}
\toprule
\textbf{Model} & \textbf{Identifier} & \textbf{Learning Rate} & \textbf{Weight Decay} & \textbf{Warmup Ratio} \\
\midrule
BERT-base & google-bert/bert-base-uncased & 2e-5 & 0.1  & 0.06 \\
BERT-large & google-bert/bert-large-uncased & 2e-5 & 0.01 & 0.1 \\
RoBERTa-large & FacebookAI/roberta-large & 1e-5 & 0.03 & 0.06 \\
RoBERTa-base & FacebookAI/roberta-base & 1e-5 & 0.05 & 0.1 \\
DeBERTa-v3-large & microsoft/deberta-v3-large & 1e-5 & 0.03 & 0.1 \\
DeBERTa-v3-base & microsoft/deberta-v3-base & 4e-5 & 0.05 & 0.1 \\
DistilBERT-base & distilbert/distilbert-base-uncased & 2e-5 & 0.03 & 0.1 \\
\bottomrule
\end{tabular}
\caption{Optimal hyperparameters for each model in multilabel malicious intent classification.}
\label{tab:hyperparams_multilabel}
\end{table*}

\section{Overview of LLMs from Experiments}
\label{appendix:llms_details}

\begin{table*}[ht!]
\scriptsize
\centering
\begin{tabular}{lllll}
\toprule
 \textbf{API Model Name} & \textbf{Knowledge Cutoff Date} & \textbf{Access Details} & \textbf{License} & \textbf{Model Size}\\\midrule
 \texttt{gpt-4o-mini} &October 2023 & OpenAI API 07.2025 & Commercial & Not Disclosed\\
 \texttt{gemini-2.0-flash} & June 2024 & Google API 07.2025 & Commercial & Not Disclosed\\ 
 \texttt{gpt-4.1-mini-2025-04-14} & June 2024 & OpenAI API 07.2025 & Commercial & Not Disclosed\\ 
 \texttt{meta-llama/Llama-3.3-70B-Instruct} &  December 2023 & DeepInfra API 07.2025 & Meta Llama 3 Community & 70B \\ 
 \texttt{google/gemma-3-27b-it} & August 2024 & DeepInfra API 07.2025 & Gemma Terms of Use & 27B \\ 
\bottomrule
\end{tabular}
\caption{Large Language Models used in our experiments.}
\label{tab:llm-models}
\end{table*}

In our experiments, we used five different cutting-edge LLMs: \textit{GPT 4o Mini},
\textit{GPT 4.1 Mini}, \textit{Gemini 2.0 Flash}, \textit{Gemma 3 27b it}, \textit{Llama 3.3 70B}. We aimed to include widely recognized, state-of-the-art models from the largest available while ensuring they remain affordable. We also selected two open-weight models to demonstrate that intent-based reasoning can be applied without access to closed models through APIs. 
 
 Table \ref{tab:llm-models} lists the Large Language Models used in our experiments, detailing their knowledge cutoff dates, access methods, licenses, and sizes. To enable evaluation on both prior and post cutoff content, we used a knowledge cutoff date of September 2024 to split the EUDisinfo and MALINT datasets accordingly. All other datasets contain only texts published prior to this date.

\section{Prompts for Malicious Intent Classification Tasks}
\label{appendix:prompts_malint}

\subsection{Binary Detection Per Each Class.}
To perform binary detection of specific malicious intent categories, we designed a prompt that conditions the LLM on a single target intent and asks for a strict \texttt{Yes}/\texttt{No} decision. The model is instructed to be conservative and only respond \texttt{Yes} when confident (see Figure~\ref{fig:prompt_binary_llms}).

\begin{figure*}[ht]
  \centering
  \begin{tcolorbox}[colback=orange!5!white, colframe=orange!50!black, 
    width=\textwidth, boxrule=0.5pt, arc=3pt, auto outer arc,
    fonttitle=\bfseries, title=Binary Detection Per Each Class]
\scriptsize
\textbf{System Prompt:} \\
 You are an assistant who detects malicious intention in text. Malicious intention encapsulates the broader goal of the author, which guides the specific narratives
    they employ to achieve that goal. Your expertise lies in detecting one malicious intention: \\
    \textit{<Here name of the malicious intent category [shortcut]>} \\
\textbf{User Prompt:} \\
Analyze the text and decide if the text contains any malicious intention: Undermining the credibility of public institutions [UCPI]
    You are very conservative in your final decisions and when you are not fully sure you answer No. \\
    Give your answer in the form of dictionary: \\
\texttt{\{} \\
\hspace*{1.5em}\texttt{"<[shortcut]>"}: \texttt{"Your answer if text include \textit{<Here name of the malicious intent category>} intent. Use only Yes or No"} \\
\texttt{\}} \\
Text: \textit{<Text to analyze>}
\medskip
  \end{tcolorbox}
  \caption{Prompt template used for binary classification of malicious intent categories with LLMs. In each instance, placeholders \textit{<Here name of the malicious intent category>} and \textit{<[shortcut]>} were replaced with one of the following categories and their respective abbreviations: Undermining the Credibility of Public Institutions [UCPI], Changing Political Views [CPV], Undermining International Organizations and Alliances [UIOA], Promoting Social Stereotypes/Antagonisms [PSSA], and Promoting Anti-scientific Views [PASV].}
  \label{fig:prompt_binary_llms}
\end{figure*}

\subsection{Multilabel Detection.}

For the multilabel setting, we use a single prompt that asks the model to evaluate the presence of all five malicious intent categories simultaneously. The model provides a \texttt{Yes}/\texttt{No} decision for each category independently and is instructed to be conservative in its judgments (see Figure~\ref{fig:prompt_multilabel_llms}).

\begin{figure*}[ht]
  \centering
  \begin{tcolorbox}[colback=orange!5!white, colframe=orange!50!black, 
    width=\textwidth, boxrule=0.5pt, arc=3pt, auto outer arc,
    fonttitle=\bfseries, title=Multilabel Detection]
\scriptsize
\textbf{System Prompt:} \\
You are an assistant who detects malicious intention in text. Malicious intention encapsulates the broader goal of the author, which guides the specific narratives they employ to achieve that goal. Your expertise lies in detecting five different malicious intentions:
\begin{enumerate}[nosep, leftmargin=*]
    \item Undermining the credibility of public institutions [UCPI]
    \item Changing political views [CPV]
    \item Undermining international organizations and alliances [UIOA]
    \item Promoting social stereotypes/antagonisms [PSSA]
    \item Promoting anti-scientific views [PASV]
\end{enumerate}
\textbf{User Prompt:} \\
 Analyze the text and decide if the text contains any malicious intentions from the following:  \\
\hspace*{1em}Undermining the credibility of public institutions [UCPI],  \\
\hspace*{1em}Changing political views [CPV], \\
\hspace*{1em}Undermining international organizations and alliances [UIOA], \\
\hspace*{1em}Promoting social stereotypes/antagonisms [PSSA], \\
\hspace*{1em}Promoting anti-scientific views [PASV]. \\[0.5ex]
You are very conservative in your final decisions and when you are not fully sure you answer No. 
 \texttt{No}. \\[0.5ex]
Give your answer in the form of dictionary: \\
\texttt{\{} \\
\hspace*{1.5em}\texttt{"UCPI"}: \texttt{"Your answer if text include Undermining the credibility of public institutions intent. Use only Yes or No"}, \\
\hspace*{1.5em}\texttt{"CPV"}: \texttt{"Your answer if text include Changing political views intent. Use only Yes or No"}, \\
\hspace*{1.5em}\texttt{"UIOA"}: \texttt{"Your answer if text include Undermining international organizations and alliances intent. Use only Yes or No"}, \\
\hspace*{1.5em}\texttt{"PSSA"}: \texttt{"Your answer if text include Promoting social stereotypes/antagonisms intent. Use only Yes or No"}, \\
\hspace*{1.5em}\texttt{"PASV"}: \texttt{"Your answer if text include Promoting anti-scientific views intent. Use only Yes or No"} \\
\texttt{\}}

Text: \textit{<Text to analyze>}

\medskip

  \end{tcolorbox}
  \caption{Prompt used for multilabel classification of malicious intent with LLMs. The system is instructed to detect five predefined categories of malicious intent within a given text. The model evaluates all categories simultaneously and returns a dictionary of binary \texttt{Yes}/\texttt{No} decisions for each. The prompt emphasizes a conservative decision-making policy: the model is instructed to respond \texttt{Yes} only when confident.}
  \label{fig:prompt_multilabel_llms}
\end{figure*}

\section{Baseline Methods and Prompts used for Intent-Based Inoculation Experiments}
\label{appendix:lucas_methods}

\subsection{Baseline Methods}

For the disinformation detection stage, we selected three strong baseline methods identified by \citet{lucas2023fighting} as top performers, particularly on human-annotated datasets such as CoAID \cite{cui2020coaid} and FakeNewsNet \cite{shu2020fakenewsnet}. The selected methods are as follows:
\begin{itemize}[nosep, leftmargin=*]
\item \textit{VaN} – A basic prompt that provides minimal, direct instructions to the LLM, serving as a foundational baseline \cite{lucas2023fighting}.
\item \textit{Z-CoT} – Builds on \textit{VaN} by encouraging step-by-step reasoning, following the zero-shot chain-of-thought prompting strategy introduced by \citet{kojima2022large}.
\item \textit{DeF-SpeC} – A more advanced prompt designed to elicit contextual, deductive, and abductive reasoning \cite{lucas2023fighting}, addressing limitations in LLMs’ ability to perform multi-step or inductive inference \cite{bang2023multitask}.
\end{itemize}

\begin{figure*}[ht] 
    \centering
    \includegraphics[width=1\textwidth]{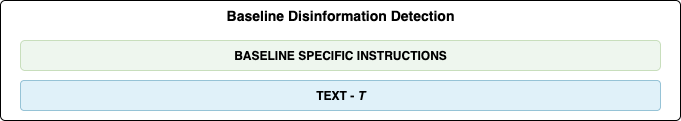} 
    \caption{The prompt template for each baseline method in disinformation detection, namely, \textit{VaN}, \textit{Z-CoT}, and \textit{DeF-SpeC}. Each baseline method differs in the \textit{Baseline Specific Instructions} block. Generally, it provides method-specific guidelines defining the task and requests for structured output. The text \(T\) represents the content passed for disinformation evaluation.}
    \label{fig:base_disinformation_detection_prompt}
\end{figure*}

We used these methods as baselines to evaluate the effectiveness of our intent-based inoculation (IBI) approach (Figure \ref{fig:base_disinformation_detection_prompt} illustrates the baseline prompt template). To adapt them to our setting, we modified the original prompts to incorporate malicious intent analysis, as introduced in the first stage of our pipeline (Section~\ref{subsec:ibi_design}). This allowed us to examine whether IBI is robust across different prompting strategies or sensitive to prompt.

Our evaluation was conducted on five datasets spanning multiple domains and genres, including news and social media. This diversity ensures a broad test of IBI’s generalizability and enables a meaningful comparison between standard baselines and our intent-augmented reasoning framework.

\subsection{Prompts used for IBI Experiments}
\label{sec:prompts}

In this section, we outline the prompt design used in our study of intent-based reasoning for disinformation detection and present templates corresponding to each stage of the IBI experiment. Due to the number and length of the prompts, we do not reproduce them in full here. The complete set of prompts is available in our online repository.

\paragraph{Baselines}
\label{sec:prompt_baseline}

Figure~\ref{fig:base_disinformation_detection_prompt} presents the baseline prompt template used for zero-shot disinformation detection. We focus on three methods introduced by \citet{lucas2023fighting}: \textit{VaN}, \textit{Z-CoT}, and \textit{DeF-SpeC}, which were selected based on their strong performance on human-annotated data. While \citet{lucas2023fighting} conducted a comprehensive evaluation across both human-annotated and LLM-generated datasets, our study considers only human-annotated examples. Accordingly, we include the top-performing methods in this setting.

\paragraph{Intent Analysis}
\label{sec:prompt_intent_analysis}

\begin{figure*}[ht] 
    \centering
    \includegraphics[width=1\textwidth]{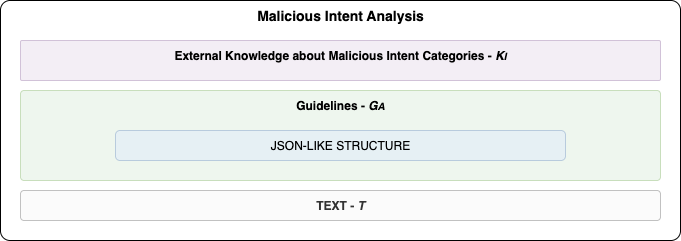}  
\caption{The prompt template for first stage of IBI experiment, namely for intent analysis. The component  \( K_I \) encapsulates knowledge about a predefined set of  malicious intent categories. Guidelines \( G_A\) determine the task and specify the structure of the expected response. Finally, the text \(T\) represents the content passed for intent analysis.}
    \label{fig:intent_detection_prompt}
\end{figure*}

Figure~\ref{fig:intent_detection_prompt} shows the final prompt template used in the first stage of the IBI experiment, which focuses on identifying malicious intent. The prompt integrates the category names and definitions from our intent taxonomy to guide model reasoning. For transparency and reproducibility, we release the full set of final prompts in our public repository.

\paragraph{Disinformation Detection with IBI}
\label{sec:prompt_ibi_final}

\begin{figure*}[ht]  
    \centering
    \includegraphics[width=1\textwidth]{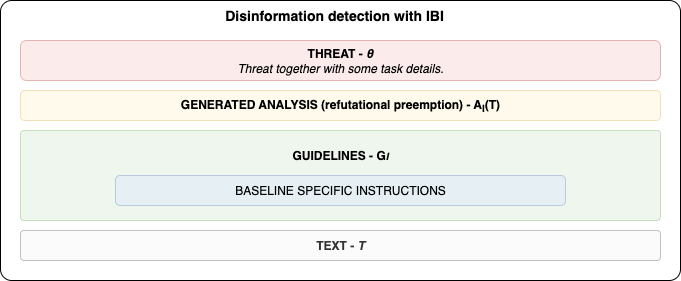}  
\caption{The prompt template for second final stage of IBI experiments, namely for disinformation detection step. The component \( \theta \) provides threat against malicious intents and gives some task details. Next component is the generated analysis \( A_I(T) \) from the output of first step of IBI experiment and finally, the text \(T\) represents the content passed for disinformation evaluation. \( G_I \) fully determine the task and specify the structure of the expected response. The \textit{Baseline Specific Instructions} block is a part of guidelines and includes different instructions depending on which baseline method was adapted to IBI experiment, namely it can be instruction from \textit{VaN}, \textit{Z-CoT}, or \textit{DeF-SpeC}}
    \label{fig:disinformation_detection_prompt}
\end{figure*}

Figure~\ref{fig:disinformation_detection_prompt} presents the final prompt template used in the second stage of the IBI experiment, which targets disinformation detection. This prompt builds on the malicious intent analysis produced in the first stage. For each test set, we evaluated three adapted prompt variants based on the \textit{VaN}, \textit{Z-CoT}, and \textit{DeF-SpeC} methods introduced by \citet{lucas2023fighting}. These adaptations align the original methods with our IBI framework. To assess their effectiveness, we compare the adapted methods against their original counterparts as baselines.

\section{McNemar’s Test for Intent-Based Inoculation Experiments}
\label{appendix:mcnemar}

To evaluate the statistical significance of intent-base inoculation, we conducted McNemar's test comparing each prompting method to its IBI-adjusted counterpart across various language models. The results, presented in Table~\ref{tab:mc_nemars_test_int}, show that IBI improves performance primarily at the significance level of 0.01 across all models and methods in the overall evaluation. However, certain cases, such as experiments on twitter posts exhibit significance for each adjusted prompting method only on GPT-4.1 Mini model. Overall, siginificance on 0.05 level or higher can be observed on about 79\%  different scenarios.

\begin{table*}[ht]
\centering
\small
\begin{tabular}{llccccc}
\toprule
Method & Split & Gemini 2.0 Flash & GPT 4.1 Mini & GPT 4o Mini & Llama3.3 70B & Gemma 3 27B \\
\midrule
 VaN       & Overall      & 0.010 & 0.010 & 0.010 & 0.050 & 0.010 \\
 VaN       & News Article & 0.010 & 0.010 & 0.010 & 0.010 & 0.010 \\
VaN       & Twitter Post & N.S.  & 0.010 & 0.010 & N.S.  & N.S.  \\
 VaN       & Prior Cutoff & 0.010 & 0.010 & 0.010 & N.S.  & 0.010 \\
 VaN       & Post Cutoff  & 0.010 & 0.010 & 0.010 & 0.050 & N.S.  \\
 Z-CoT     & Overall      & 0.010 & 0.010 & 0.010 & 0.010 & 0.010 \\
Z-CoT     & News Article & 0.010 & 0.010 & 0.010 & 0.010 & 0.010 \\
 Z-CoT     & Twitter Post & N.S.  & 0.010 & 0.010 & N.S.  & 0.010 \\
 Z-CoT     & Prior Cutoff & 0.010 & 0.010 & 0.010 & 0.010 & 0.010 \\
 Z-CoT     & Post Cutoff  & 0.010 & 0.010 & N.S.  & 0.010 & N.S.  \\
DeF\_Spec & Overall      & 0.010 & 0.010 & 0.010 & 0.010 & 0.010 \\
 DeF\_Spec & News Article & 0.010 & 0.010 & 0.050 & 0.010 & 0.010 \\
DeF\_Spec & Twitter Post & N.S.  & 0.010 & 0.010 & N.S.  & N.S.  \\
 DeF\_Spec & Prior Cutoff & 0.010 & 0.010 & 0.010 & 0.010 & 0.010 \\
 DeF\_Spec & Post Cutoff  & N.S.  & N.S.  & N.S.  & 0.010 & N.S.  \\
\bottomrule
\end{tabular}
     \caption{Results of McNemar's test, comparing each prompting method (\textit{VaN}, \textit{Z-CoT}, and \textit{DeF-Spec}) against its IBI-adjusted counterpart across various language models. The values represent significance levels for different evaluation metrics, with \textit{N.S} as \textit{Non-Significant} indicating no statistically significant difference at the 0.05 threshold.}
\label{tab:mc_nemars_test_int}
\end{table*}

\section{Intent-Based Inoculation Experiments Results on Different Datasets}
\label{appendix:ibi_results_specific_datasets}

Table~\ref{tab:ibi_results_all_datasets} reports F\textsubscript{1} scores across four disinformation datasets for various models and prompting strategies, further enhanced with intent-based inoculation (IBI). The table compares the Base setup with IBI, showing that IBI consistently improves performance across nearly all models and datasets. The largest gains are observed on datasets containing longer-form news articles, while improvements on ECTF are more modest, likely due to limited context in shorter social media posts. Overall, these results demonstrate that intent-augmented reasoning effectively enhances disinformation detection, particularly for longer-form texts, across diverse datasets and model architectures.

\begin{table*}[!ht]
\footnotesize
\centering
\begin{tabular}{lcccccccccc}
\toprule
 & \multicolumn{2}{c}{\textit{GPT 4o Mini}} 
 & \multicolumn{2}{c}{\textit{GPT 4.1 Mini}} 
 & \multicolumn{2}{c}{\textit{Gemini 2.0 Flash}} 
 & \multicolumn{2}{c}{\textit{Gemma 3 27b it}} 
 & \multicolumn{2}{c}{\textit{Llama 3.3 70B}} \\
\cmidrule(lr){2-3} \cmidrule(lr){4-5} \cmidrule(lr){6-7} \cmidrule(lr){8-9} \cmidrule(lr){10-11}
 & Base & IBI & Base & IBI & Base & IBI & Base & IBI & Base & IBI \\
\midrule
\multicolumn{11}{l}{\textbf{CoAID}} \\
VaN       & 0.531 & 0.627 & 0.480 & 0.599 & 0.631 & 0.607 & 0.618 & 0.650 & 0.654 & 0.680 \\
Z-CoT     & 0.532 & 0.628 & 0.468 & 0.611 & 0.588 & 0.603 & 0.388 & 0.660 & 0.628 & 0.699 \\
DeF\_Spec & 0.507 & 0.566 & 0.496 & 0.624 & 0.574 & 0.610 & 0.617 & 0.651 & 0.582 & 0.646 \\
\midrule
\multicolumn{11}{l}{\textbf{ECTF}} \\
VaN       & 0.812 & 0.821 & 0.772 & 0.758 & 0.743 & 0.713 & 0.769 & 0.733 & 0.799 & 0.732 \\
Z-CoT     & 0.796 & 0.830 & 0.766 & 0.762 & 0.728 & 0.703 & 0.589 & 0.728 & 0.789 & 0.745 \\
DeF\_Spec & 0.783 & 0.837 & 0.790 & 0.784 & 0.801 & 0.812 & 0.800 & 0.755 & 0.773 & 0.736 \\
\midrule
\multicolumn{11}{l}{\textbf{ISOTFakeNews}} \\
VaN       & 0.776 & 0.889 & 0.681 & 0.665 & 0.650 & 0.761 & 0.529 & 0.762 & 0.701 & 0.720 \\
Z-CoT     & 0.761 & 0.890 & 0.620 & 0.662 & 0.591 & 0.678 & 0.514 & 0.751 & 0.683 & 0.726 \\
DeF\_Spec & 0.741 & 0.782 & 0.780 & 0.744 & 0.780 & 0.832 & 0.634 & 0.790 & 0.623 & 0.686 \\
\midrule
\multicolumn{11}{l}{\textbf{EUDisinfo}} \\
VaN       & 0.682 & 0.860 & 0.678 & 0.856 & 0.693 & 0.842 & 0.821 & 0.873 & 0.784 & 0.866 \\
Z-CoT     & 0.727 & 0.847 & 0.653 & 0.845 & 0.705 & 0.846 & 0.802 & 0.877 & 0.765 & 0.885 \\
DeF\_Spec & 0.789 & 0.849 & 0.755 & 0.846 & 0.794 & 0.829 & 0.867 & 0.885 & 0.803 & 0.886 \\
\bottomrule
\end{tabular}
\caption{F\textsubscript{1} scores on four disinformation datasets for competitive prompting methods and their enhancement with Intent-Based Inoculation.}
\label{tab:ibi_results_all_datasets}
\end{table*}

\section{Impact of Intent Prediction on Disinformation Detection with IBI}
\label{sec:ablation_impact_intent_on_ibi}

Table~\ref{tab:average_intent_influence} shows the average F\textsubscript{1} scores for disinformation detection, split by whether the predicted intent matches the gold labels (“Correct”) or not (“Incorrect”). Overall, accurate intent predictions generally improve disinformation detection performance, as evidenced by higher F\textsubscript{1} scores in the “Correct” subsets for most intents and models. Notable exceptions, such as UIOA for some models, indicate that even mispredicted intents can still provide useful signals for detection. While the analysis evaluates each intent independently, the underlying multi-label nature of intent prediction means correct predictions in other intents could offset errors, making the actual impact of intent on F\textsubscript{1} more nuanced. These results highlight that intent-augmented reasoning can enhance detection, but the relationship between intent accuracy and disinformation detection is complex and context-dependent. 

Table~\ref{tab:average_intent_influence} presents average results across all three competitive baseline methods improved to intent-based resoning. In contrast, Tables~\ref{tab:van_intent_influence}, \ref{tab:zcot_intent_influence}, and \ref{tab:defspec_intent_influence} report results for each specific baseline method further enhanced with intent-based reasoning.

\begin{table*}[h!]
\centering
\scriptsize
\setlength{\tabcolsep}{5pt}
\begin{tabular}{lcccccccccc}
\toprule
\multirow{2}{*}{Model} & \multicolumn{2}{c}{CPV} & \multicolumn{2}{c}{PSSA} & \multicolumn{2}{c}{UIOA} & \multicolumn{2}{c}{PASV} & \multicolumn{2}{c}{UCPI} \\
\cmidrule(lr){2-3} \cmidrule(lr){4-5} \cmidrule(lr){6-7} \cmidrule(lr){8-9} \cmidrule(lr){10-11}
 & Correct & Incorrect & Correct & Incorrect & Correct & Incorrect & Correct & Incorrect & Correct & Incorrect \\
\midrule
gpt-4o-mini & 0.914 & 0.787 & 0.882 & 0.809 & 0.860 & 0.864 & 0.866 & 0.850 & 0.905 & 0.796 \\
Llama-3.3-70B-Instruct & 0.889 & 0.825 & 0.876 & 0.839 & 0.856 & 0.899 & 0.862 & 0.888 & 0.836 & 0.941 \\
gpt-4.1-mini & 0.872 & 0.869 & 0.886 & 0.831 & 0.862 & 0.908 & 0.862 & 0.905 & 0.876 & 0.863 \\
gemini-2.0-flash & 0.878 & 0.821 & 0.855 & 0.866 & 0.845 & 0.902 & 0.854 & 0.871 & 0.865 & 0.842 \\
google/gemma-3-27b-it & 0.853 & 0.806 & 0.834 & 0.799 & 0.824 & 0.824 & 0.815 & 0.862 & 0.876 & 0.722 \\
\bottomrule
\end{tabular}
\caption{Average F1 scores for disinformation detection across three methods (VaN, Z-CoT, DeF-Spec), split by correct and incorrect intent prediction for each intent.}
\label{tab:average_intent_influence}
\end{table*}

\begin{table*}[!ht]
\scriptsize
\centering
\begin{tabular}{lcc|cc|cc|cc|cc}
\toprule
 & \multicolumn{2}{c}{CPV} & \multicolumn{2}{c}{PSSA} & \multicolumn{2}{c}{UIOA} & \multicolumn{2}{c}{PASV} & \multicolumn{2}{c}{UCPI} \\
\cmidrule(lr){2-3} \cmidrule(lr){4-5} \cmidrule(lr){6-7} \cmidrule(lr){8-9} \cmidrule(lr){10-11}
Model & Correct & Incorrect & Correct & Incorrect & Correct & Incorrect & Correct & Incorrect & Correct & Incorrect \\
\midrule
gpt-4o-mini & 0.910 & 0.783 & 0.883 & 0.792 & 0.852 & 0.867 & 0.860 & 0.848 & 0.903 & 0.788 \\
Llama-3.3-70B & 0.881 & 0.829 & 0.870 & 0.843 & 0.855 & 0.889 & 0.857 & 0.889 & 0.834 & 0.933 \\
gpt-4.1-mini & 0.870 & 0.879 & 0.882 & 0.850 & 0.864 & 0.912 & 0.861 & 0.921 & 0.871 & 0.877 \\
gemini-2.0-flash & 0.869 & 0.830 & 0.849 & 0.871 & 0.843 & 0.896 & 0.856 & 0.853 & 0.871 & 0.821 \\
gemma-3-27b-it & 0.853 & 0.800 & 0.831 & 0.792 & 0.828 & 0.800 & 0.812 & 0.857 & 0.879 & 0.707 \\
\bottomrule
\end{tabular}
\caption{F\textsubscript{1} scores with VaN + IBI, split by intent prediction correctness.}
\label{tab:van_intent_influence}
\end{table*}

\begin{table*}[!ht]
\scriptsize
\centering
\begin{tabular}{lcc|cc|cc|cc|cc}
\toprule
 & \multicolumn{2}{c}{CPV} & \multicolumn{2}{c}{PSSA} & \multicolumn{2}{c}{UIOA} & \multicolumn{2}{c}{PASV} & \multicolumn{2}{c}{UCPI} \\
\cmidrule(lr){2-3} \cmidrule(lr){4-5} \cmidrule(lr){6-7} \cmidrule(lr){8-9} \cmidrule(lr){10-11}
Model & Correct & Incorrect & Correct & Incorrect & Correct & Incorrect & Correct & Incorrect & Correct & Incorrect \\
\midrule
gpt-4o-mini & 0.911 & 0.766 & 0.872 & 0.796 & 0.850 & 0.847 & 0.857 & 0.832 & 0.913 & 0.759 \\
Llama-3.3-70B & 0.891 & 0.818 & 0.873 & 0.843 & 0.854 & 0.904 & 0.860 & 0.889 & 0.833 & 0.945 \\
gpt-4.1-mini & 0.862 & 0.860 & 0.874 & 0.825 & 0.844 & 0.931 & 0.854 & 0.889 & 0.865 & 0.855 \\
gemini-2.0-flash & 0.847 & 0.819 & 0.833 & 0.847 & 0.825 & 0.879 & 0.828 & 0.866 & 0.840 & 0.830 \\
gemma-3-27b-it & 0.837 & 0.786 & 0.818 & 0.774 & 0.805 & 0.809 & 0.794 & 0.857 & 0.858 & 0.707 \\
\bottomrule
\end{tabular}
\caption{F\textsubscript{1} scores with Z-CoT + IBI, split by intent prediction correctness.}
\label{tab:zcot_intent_influence}
\end{table*}

\begin{table*}[!ht]
\scriptsize
\centering
\begin{tabular}{lcc|cc|cc|cc|cc}
\toprule
 & \multicolumn{2}{c}{CPV} & \multicolumn{2}{c}{PSSA} & \multicolumn{2}{c}{UIOA} & \multicolumn{2}{c}{PASV} & \multicolumn{2}{c}{UCPI} \\
\cmidrule(lr){2-3} \cmidrule(lr){4-5} \cmidrule(lr){6-7} \cmidrule(lr){8-9} \cmidrule(lr){10-11}
Model & Correct & Incorrect & Correct & Incorrect & Correct & Incorrect & Correct & Incorrect & Correct & Incorrect \\
\midrule
gpt-4o-mini & 0.921 & 0.812 & 0.892 & 0.839 & 0.876 & 0.878 & 0.880 & 0.869 & 0.899 & 0.841 \\
Llama-3.3-70B & 0.894 & 0.829 & 0.886 & 0.829 & 0.861 & 0.904 & 0.868 & 0.885 & 0.840 & 0.945 \\
gpt-4.1-mini & 0.885 & 0.867 & 0.901 & 0.819 & 0.878 & 0.881 & 0.871 & 0.906 & 0.892 & 0.857 \\
gemini-2.0-flash & 0.918 & 0.814 & 0.882 & 0.879 & 0.866 & 0.932 & 0.877 & 0.896 & 0.885 & 0.874 \\
gemma-3-27b-it & 0.868 & 0.832 & 0.852 & 0.830 & 0.840 & 0.864 & 0.840 & 0.871 & 0.892 & 0.752 \\
\bottomrule
\end{tabular}
\caption{F\textsubscript{1} scores with DeF-Spec + IBI, split by intent prediction correctness.}
\label{tab:defspec_intent_influence}
\end{table*}

\section{Cross-lingual results with IBI}
\label{appendix:cross-lingual-res-ibi} 

\begin{table*}[h!]
\centering
\scriptsize
\renewcommand{\arraystretch}{1.1}
\begin{tabular}{lcc}
\hline
\textbf{Language} & \textbf{Base} & \textbf{IBI} \\
\hline
German   & 0.794 $\pm$ 0.06 & 0.911 $\pm$ 0.02 \\
Spanish  & 0.683 $\pm$ 0.08 & 0.828 $\pm$ 0.04\\
Estonian & 0.716 $\pm$ 0.11 & 0.892 $\pm$ 0.04\\
French   & 0.611 $\pm$ 0.06& 0.749 $\pm$ 0.05\\
Polish   & 0.709 $\pm$ 0.1 & 0.846 $\pm$ 0.05\\
Russian  & 0.619 $\pm$ 0.07 & 0.735 $\pm$ 0.02\\
\hline
\end{tabular}
\caption{Comparison of Base vs IBI performance across languages. Results present averaged F\textsubscript{1} scores together with standard deviations over all models and methods.}
\label{tab:ibi_vs_base_per_language_avg_std}
\end{table*}

Table~\ref{tab:ibi_f1_languages_full} presents F\textsubscript{1} scores computed on approximately 500 test texts per language across six languages in the EUvsDisinfo dataset. Overall, IBI consistently improves performance across all models and languages, demonstrating the effectiveness of intent-augmented reasoning in disinformation detection. These results confirm that incorporating intent information provides robust performance gains even for low-resource languages like Estonian.

Table~\ref{tab:ibi_vs_base_per_language_avg_std} shows that the IBI approach consistently outperforms the Base model across all languages, achieving higher averaged F\textsubscript{1} scores with generally lower variability. The reported values are averaged over all models and methods and are presented together with their standard deviations, providing a clear indication of performance stability.

\begin{table*}[!ht]
\scriptsize
\centering
\begin{tabular}{lcccccc}
\toprule
 & \multicolumn{2}{c}{\textbf{Spanish}} & \multicolumn{2}{c}{\textbf{German}} & \multicolumn{2}{c}{\textbf{Polish}} \\
\cmidrule(lr){2-3} \cmidrule(lr){4-5} \cmidrule(lr){6-7}
 & Base & IBI & Base & IBI & Base & IBI \\
\midrule
\multicolumn{7}{l}{\textit{GPT 4o Mini}}\\
VaN & 0.715 & \perc{0.715}{0.868} & 0.806 & \perc{0.806}{0.925} & 0.734 & \perc{0.734}{0.842} \\
Z-CoT & 0.710 & \perc{0.710}{0.877} & 0.813 & \perc{0.813}{0.932} & 0.727 & \perc{0.727}{0.840} \\
DeF-Spec & 0.742 & \perc{0.742}{0.825} & 0.850 & \perc{0.850}{0.913} & 0.696 & \perc{0.696}{0.775} \\
\midrule
\multicolumn{7}{l}{\textit{Llama 3.3 70B}}\\
VaN & 0.719 & \perc{0.719}{0.817} & 0.813 & \perc{0.813}{0.892} & 0.809 & \perc{0.809}{0.847} \\
Z-CoT & 0.710 & \perc{0.710}{0.820} & 0.814 & \perc{0.814}{0.897} & 0.733 & \perc{0.733}{0.844} \\
DeF-Spec & 0.738 & \perc{0.738}{0.805} & 0.833 & \perc{0.833}{0.909} & 0.860 & \perc{0.860}{0.855} \\
\midrule
\multicolumn{7}{l}{\textit{GPT 4.1 Mini}}\\
VaN & 0.561 & \perc{0.561}{0.775} & 0.693 & \perc{0.693}{0.887} & 0.579 & \perc{0.579}{0.775} \\
Z-CoT & 0.549 & \perc{0.549}{0.775} & 0.671 & \perc{0.671}{0.882} & 0.546 & \perc{0.546}{0.775} \\
DeF-Spec & 0.717 & \perc{0.717}{0.796} & 0.824 & \perc{0.824}{0.895} & 0.690 & \perc{0.690}{0.796} \\
\midrule
\multicolumn{7}{l}{\textit{Gemini 2.0 Flash}}\\
VaN & 0.560 & \perc{0.560}{0.797} & 0.737 & \perc{0.737}{0.885} & 0.556 & \perc{0.556}{0.873} \\
Z-CoT & 0.568 & \perc{0.568}{0.791} & 0.756 & \perc{0.756}{0.886} & 0.611 & \perc{0.611}{0.851} \\
DeF-Spec & 0.744 & \perc{0.744}{0.801} & 0.845 & \perc{0.845}{0.926} & 0.802 & \perc{0.802}{0.904} \\
\midrule
\multicolumn{7}{l}{\textit{Google Gemma 3-27B}}\\
VaN & 0.711 & \perc{0.711}{0.871} & 0.792 & \perc{0.792}{0.941} & 0.809 & \perc{0.809}{0.903} \\
Z-CoT & 0.722 & \perc{0.722}{0.869} & 0.814 & \perc{0.814}{0.942} & 0.813 & \perc{0.813}{0.901} \\
DeF-Spec & 0.782 & \perc{0.782}{0.880} & 0.846 & \perc{0.846}{0.953} & 0.860 & \perc{0.860}{0.912} \\
\bottomrule
 & \multicolumn{2}{c}{\textbf{French}} & \multicolumn{2}{c}{\textbf{Russian}} & \multicolumn{2}{c}{\textbf{Estonian}} \\
\cmidrule(lr){2-3} \cmidrule(lr){4-5} \cmidrule(lr){6-7}
 & Base & IBI & Base & IBI & Base & IBI \\
\midrule
\multicolumn{7}{l}{\textit{GPT 4o Mini}}\\
VaN & 0.618 & \perc{0.618}{0.757} & 0.644 & \perc{0.644}{0.749} & 0.732 & \perc{0.732}{0.945} \\
Z-CoT & 0.632 & \perc{0.632}{0.759} & 0.657 & \perc{0.657}{0.750} & 0.727 & \perc{0.727}{0.933} \\
DeF-Spec & 0.684 & \perc{0.684}{0.739} & 0.711 & \perc{0.711}{0.738} & 0.705 & \perc{0.705}{0.859} \\
\midrule
\multicolumn{7}{l}{\textit{Llama 3.3 70B}}\\
VaN & 0.629 & \perc{0.629}{0.773} & 0.626 & \perc{0.626}{0.745} & 0.756 & \perc{0.756}{0.902} \\
Z-CoT & 0.628 & \perc{0.628}{0.755} & 0.643 & \perc{0.643}{0.752} & 0.751 & \perc{0.751}{0.907} \\
DeF-Spec & 0.661 & \perc{0.661}{0.763} & 0.664 & \perc{0.664}{0.764} & 0.805 & \perc{0.805}{0.908} \\
\midrule
\multicolumn{7}{l}{\textit{GPT 4.1 Mini}}\\
VaN & 0.545 & \perc{0.545}{0.676} & 0.509 & \perc{0.509}{0.718} & 0.546 & \perc{0.546}{0.835} \\
Z-CoT & 0.533 & \perc{0.533}{0.676} & 0.505 & \perc{0.505}{0.730} & 0.551 & \perc{0.551}{0.833} \\
DeF-Spec & 0.648 & \perc{0.648}{0.715} & 0.667 & \perc{0.667}{0.722} & 0.637 & \perc{0.637}{0.847} \\
\midrule
\multicolumn{7}{l}{\textit{Gemini 2.0 Flash}}\\
VaN & 0.502 & \perc{0.502}{0.729} & 0.489 & \perc{0.489}{0.705} & 0.579 & \perc{0.579}{0.871} \\
Z-CoT & 0.514 & \perc{0.514}{0.711} & 0.523 & \perc{0.523}{0.696} & 0.601 & \perc{0.601}{0.857} \\
DeF-Spec & 0.641 & \perc{0.641}{0.736} & 0.655 & \perc{0.655}{0.733} & 0.767 & \perc{0.767}{0.852} \\
\midrule
\multicolumn{7}{l}{\textit{Google Gemma 3-27B}}\\
VaN & 0.621 & \perc{0.621}{0.800} & 0.644 & \perc{0.644}{0.736} & 0.849 & \perc{0.849}{0.939} \\
Z-CoT & 0.626 & \perc{0.626}{0.807} & 0.655 & \perc{0.655}{0.738} & 0.857 & \perc{0.857}{0.943} \\
DeF-Spec & 0.688 & \perc{0.688}{0.839} & 0.699 & \perc{0.699}{0.744} & 0.874 & \perc{0.874}{0.952} \\
\bottomrule
\end{tabular}

\caption{F\textsubscript{1} scores on the EUvsDisinfo dataset separated by six languages computed on about 500 test texts per language. Results report F\textsubscript{1} for competitive prompting methods as the baseline (\textit{Base}) and their enhancement with intent-based reasoning (\textit{IBI}).}

\label{tab:ibi_f1_languages_full}
\end{table*}

\section{MALINT Comparison with Existing Datasets}
\label{appendix:table-comparison-benchmarks}
 
Table~\ref{tab:table_comparison_to_other_banchmarks} provides a comparison of MALINT with existing datasets and papers that contain intent analysis for disinformation. The comparison highlights several key aspects:

\begin{itemize}[nosep, leftmargin=*]
    \item \textbf{Dataset origin:} While some datasets are enhanced versions of existing collections \citet{Zhou_2022}, MALINT and \citet{modzelewski2024mipd} are fully original datasets. Moreover, one study experiments with intent behind misinformation, but does not present any novel dataset \citet{wang2024misinformation}.
    \item \textbf{Intent categorization:} MALINT introduces 6 categories focused on current global disinformation, whereas other datasets either provide binary labels or local/national intent categories.
    \item \textbf{Annotation quality:} Unlike some datasets relying partially on algorithmic or weak annotation, MALINT is fully human-annotated and additionally documents each step of the annotation process, ensuring transparency and reproducibility.
    \item \textbf{Language coverage:} MALINT is in English and covers global disinformation, expanding beyond datasets that are either language-specific or region-specific.
\end{itemize}

Overall, MALINT stands out by combining original data, comprehensive human annotation, multi-step transparency, and a focus on globally relevant disinformation intents.

\begin{table*}[ht]
\centering
\scriptsize
\begin{tabular}{@{}p{2cm}p{2.5cm}p{3.0cm}p{1.8cm}p{1cm}p{1.5cm}p{1.5cm}@{}}
\toprule
Paper & Dataset & Intent Categories & Fully Annotated & Language & Algorithmic Annotation & Annotation from Each Step \\ \midrule
\citet{Zhou_2022} & Enhanced from Existing & Intentional / Unintentional & No & English & Yes & No \\
\midrule
\citet{modzelewski2024mipd} & Original & 9 categories focused on Polish disinformation & Yes & Polish & Fully annotated by humans & No \\
\midrule
\citet{wang2024misinformation} & No dataset presented & Hierarchy of expressed intents of articles (lowest-level labels: Popularize, Clout, Smear, Conflict, Connect) & No & English & Yes & No \\
\midrule
Our MALINT & Original & 6 categories focused on current global disinformation & Yes & English & Fully annotated by humans & Yes \\ \bottomrule
\end{tabular}
\caption{Comparison of datasets related to MALINT, highlighting the novelty of the MALINT dataset}
\label{tab:table_comparison_to_other_banchmarks}
\end{table*}

\section{Generalizability of Intent Categories and IBI Experiments}

The generalizability of the intent categories used in this work was empirically validated. This validation demonstrates that the categories are applicable across multiple disinformation datasets.

Experiments with the Intent-Based Inoculation (IBI) approach were conducted on five disinformation datasets, including CoAID, ISOT Fake News, and ECTF. To demonstrate that these datasets encompass global information and context beyond Europe, topic modeling was performed using BERTopic with GPT-4o-mini.  

Example topics for three of the datasets are provided below:

\textbf{ECTF - Global Topics:}  
COVID-19 updates Canada; Sat Bhakti and Health; Israel coronavirus vaccine; Global COVID-19 Solidarity; Bioweapons in Wuhan; Wuhan virus theories; Vaccine funding initiatives; Trump and COVID-19

\textbf{ISOT Fake News - Global Topics:}  
Rohingya refugee crisis; Venezuela political crisis; Mugabe's Political Crisis; Brazilian Corruption Scandal; North Korea Sanctions; Duterte's War on Drugs; Saudi anti-corruption purge

\textbf{CoAID - Global Topics:}  
Global COVID-19 Response; Coronaviruses and Animals; Racial health disparities; Economic impact of COVID-19; COVID-19 Vaccine Distribution; Tuberculosis Awareness and Response

\end{document}